\documentclass[journal]{IEEEtran}
\usepackage{cite}

\usepackage{footmisc}

\ifCLASSINFOpdf
  \usepackage[pdftex]{graphicx}
\else
\fi
\usepackage{footmisc}
\usepackage{multirow}
\usepackage{color}
\usepackage{amsmath}
\usepackage{amssymb}
\DeclareMathOperator*{\argmax}{arg\,max}
\DeclareMathOperator*{\argmin}{arg\,min}
\begin{document}
%
\title{Reinforcement Learning and Deep Learning based Lateral Control for Autonomous Driving}
\author{
	\IEEEauthorblockN{Dong~Li, Dongbin~Zhao, Qichao~Zhang, and Yaran~Chen}
	\IEEEauthorblockA{\\The State Key Laboratory of Management and Control for Complex Systems, \\
		Institute of Automation, Chinese Academy of Sciences, Beijing, CHINA\\
	University of Chinese Academy of Sciences, Beijing, CHINA\\}
\thanks{Corresponding Author: Dongbin Zhao (Email: dongbin.zhao@ia.ac.cn)}
}
\maketitle

\begin{abstract}
This paper investigates the vision-based autonomous driving with deep learning and reinforcement learning methods. Different from the end-to-end learning method, our method breaks the vision-based lateral control system down into a perception module and a control module. The perception module which is based on a multi-task learning neural network first takes a driver-view image as its input and predicts  the track features. The control module which is based on reinforcement learning then makes a control decision based on these features. In order to improve the data efficiency, we propose visual TORCS (VTORCS), a deep reinforcement learning environment which is based on the open racing car simulator (TORCS). By means of the provided functions, one can train an agent with the input of an image or various physical sensor measurement, or evaluate the perception algorithm on this simulator. The trained reinforcement learning controller outperforms the linear quadratic regulator (LQR) controller and model predictive control (MPC) controller on different tracks. The experiments demonstrate that the perception module shows promising performance and the controller is capable of controlling the vehicle drive well along the track center with visual input.
\end{abstract}


%
\IEEEpeerreviewmaketitle

%
%
%
%
\renewcommand{\footnoterule}{%
	\kern -3pt
	\hrule width 250pt height 0.2pt
	\kern 2pt
}

\section{Introduction}
\IEEEPARstart{I}{n} recent years, artificial intelligence (AI) has flourished in many fields such as autonomous driving\cite{Maurer2016Autonomous}\cite{geiger2012are}, games\cite{mnih2015human-level}\cite{silver2017mastering}, and engineering applications \cite{en10101525}\cite{bu2016tnnls}. As one of the most popular topics, autonomous driving has drawn great attention both from the academic and industrial communities and is thought to be the next revolution in the intelligent transportation system. The autonomous driving system mainly consists of four modules: an environment perception module, a trajectory planning module, a control module, and an actuator mechanism module. The initial perception methods\cite{urmson2008autonomous}\cite{montemerlo2008junior} are based on the expensive LIDARs which usually cost tens of thousands of dollars. The high cost limits their large-scale applications to the ordinary vehicles. Recently, more attention is paid to the image-based methods\cite{zhaodeep2017} of which the core sensor, i.e. camera is relatively cheap and already equipped on most vehicles. Some of these perception methods have been developed into products\cite{Stein2006System}\cite{Stein2013Collision}.  In this paper, we focus on the lateral control problem based on the image captured by the onboard camera.

The vision-based lateral control methods can be divided into two categories: the end-to-end control methods and the perception and control separation methods. The end-to-end control system directly maps the observation to the desired output. In the context of vehicle control problem, a typical approach of the end-to-end learning is imitation learning in which a classifier\cite{Lecun2005Off} or regressor\cite{silver2008high} is learned for predicting the expert's control command when encountering the same observation. However, since the predicted action affects the following observation, a small error will accumulate and lead the learner to a totally different future observation\cite{ross2010b}. Thus the end-to-end control methods usually need a large dataset or data augmentation\cite{ross2010a} process to enhance the coverage of the observation space. Otherwise, the learner will learn a poor policy. In contrast to the end-to-end control methods, the perception and control separation methods disassemble the control pipeline into a perception module and a control module\cite{TaylorIJRR99}. The perception module takes the image as its input and extracts the features that are used to locate the vehicle on the track. The control module is responsible for making an optimal decision to follow the desired trajectory. Since the perception and control separation methods provide the flexibility that one can employ the state-of-the-art algorithm for perception and control, we design our vision-based lateral control algorithm in this framework.

The main task for the perception module is to extract useful features from the image and locate the vehicle in the track. The previous works perceive the underlying features which include lane boundaries\cite{gurghian2016deeplanes}, distance to lane boundaries\cite{chen2015deepdriving}, vehicle poses\cite{Chen7780605}, and road curvature\cite{Shen6338884}, etc. from the image. By obtaining the distance to the lane boundaries and the heading angle difference with the lane heading direction, the vehicle can be located in the track coordinate. In fact, these features show strong visual correlation. For example, the lane boundaries can present the bending degree of the track, and the curvature estimation describes the bending degree with precise value, i.e. curvature. Thus, these two perception tasks are correlated and show some common features. However, the above works solve the perception tasks separately. To utilize the general features and improve the learning performance, we formulate the perception problem in the framework of the  multi-task learning (MTL) convolutional neural network (CNN), which is able to exploit the shared information of different tasks and has been successfully applied to many fields like dangerous object detection\cite{CHEN2017}.

Another important module is the control module whose objective is to generate the optimal or near-optimal control command that keeps the vehicle follow the trajectory made by the planning module. The popular control methods include linear quadratic regulator (LQR)\cite{yi2001investigation}, fuzzy logic\cite{Wang2015Lateral}, and model predictive control (MPC)\cite{borrelli2005mpc}. However, the aforementioned control methods require system model. Since the vehicle is a strong nonlinear dynamic system running in an uncertain environment, it is hard to approximate an accurate model. Recent efforts try to employ model-free methods instead and learn from the raw sensors data\cite{lillicrap2015continuous}\cite{wang2009adaptive}. As a category of data-driven methods, reinforcement learning (RL)\cite{sutton1998reinforcement} which is inspired by the decision-making process of animals evaluates and improves its control policy by interacting with an environment. The agent is capable of learning an optimal or near-optimal policy from evaluative feedback based on the experience data\cite{zhang2016cyber}\cite{Modares2013}. Recent years have seen many exciting RL applications in the context of autonomous driving such as adaptive cruise control\cite{ZHAO201457,zhao2017model,zhu2018cacc}, lane keeping\cite{lee2017autonomous}, and obstacle avoidance\cite{Manuelli2015}. In the RL framework, the vehicle lateral control is a continuous state and continuous action decision-making problem. Therefore, we employ the policy gradient algorithm to solve this problem. For the vision-based lateral control system, first the driver-view image is fed into the perception module based on the MTL neural network to predict the key track features, then a following RL control module maps these track features to a steering action and keeps the vehicle following the trajectory. We call it the MTL-RL controller. Note the desired trajectory is set to lane center line in this paper.

\begin{figure}[!t]
	\centering
	\includegraphics[scale=0.45]{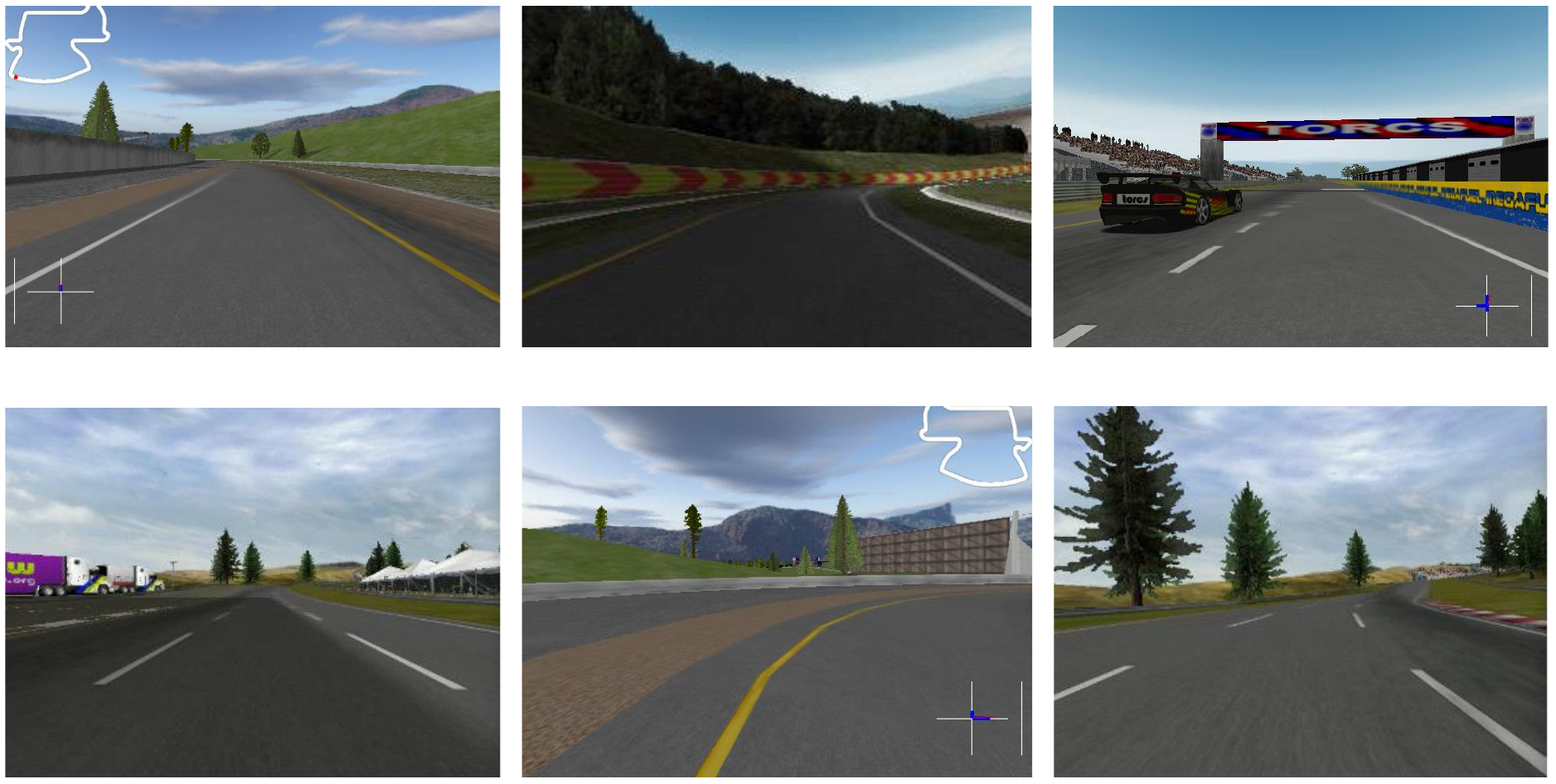}
	\caption{Traffic scenes in VTORCS with different settings. In order to implement perception and control algorithms, one can set up the VTORCS with various configurations such as different lane numbers, vehicle numbers, and track curvature etc. to obtain a desired simulation environment.}
	\label{fig_track_scene}
\end{figure}

The RL agent evaluates and improves its control policy in a trial-and-error manner which usually takes numerous samples to converge. Thus, it would be dangerous and costly to train an agent on a real vehicle. Additionally, training a deep neural network like CNN also needs a large set of samples. For the purpose of training and evaluating the perception and control algorithms, an autonomous driving environment which simultaneously integrates the image processing functions and the agent learning functions is developed. The Open Racing Car Simulator (TORCS)\cite{TORCS} is not only an ordinary vehicle racing game but also serves as an AI research platform for many years. The game engine provides high-quality traffic scenes and vehicle dynamics and kinetics models. It was the official competition software for the 2008 IEEE World Congress on Computational Intelligence (WCCI)\cite{Loiacono2008The} and the 2009 Simulated Car Racing (SCR) Championship\cite{loiacono20102009}. 

The existing TORCS releases include official release\cite{TORCS}, SCR release\cite{loiacono20102009}, and DeepDriving release\cite{chen2015deepdriving}. The official TORCS provides functions like track definition, vehicle model, and basic software framework but AI research extensions. Thus, the SCR organizers release a version with various sensors support such as range radar and speed sensor, etc. The corresponding sensor data is the low-dimensional physical measurements. It can only provide a small fixed-size image ($ 64 \times 64 $). Moreover, the employed user datagram protocol (UDP) data transmission tool is not secure because we cannot ensure all messages are received by the receiver e.g. the RL agent. Thus, this framework has limitations in image processing and communication. The DeepDriving release is specialized in image processing features like image capturing and labeling. However, it does not provide the control supports for the RL agent training and evaluating. Therefore, a new simulator which can provide perception and control supports is urgently needed. Motivated by this, we propose the visual TORCS (VTORCS), an OpenAI gym-like environment which provides multiple physical sensors, multi-view image processing, and plenty of easy-to-use RL utilities such as state and reward definition. As shown in Fig. \ref{fig_track_scene}, one can set up the VTORCS with various configurations such as different lane numbers, vehicle numbers, and track curvature etc. to obtain the desired simulation environment with different difficulty levels. At every time step, the agent can retrieve the physical or visual observations from the VTORCS and make a decision according to the underlying policy, then the action is sent back to the VTORCS where it finishes the one-step simulation.

In the presented paper, we design a vision-based lateral controller by integrating an MTL perception module and an RL control module. In the perception module, in order to address the insufficient locating precision issue which may cause vehicle out of track shown later, we analyze the correlated tasks and introduce a track orientation classification task as the auxiliary task. In the RL controller module, a reward function based on the geometrical relationship is designed to guide the policy learning. Additionally, an autonomous driving simulator VTORCS is developed for the general algorithm implementation. The experiment results\footnote{Video material provided: https://github.com/dongleecsu/VisualTORCS} validate the promising performance of these two modules and the effectiveness of the vision-based lateral controller.

The remainder of this paper is organized as follows. In section II, we introduce the VTORCS environment in details. In section III, we define the vision-based lateral control problem and introduce the background of RL and MTL. We give an overview of the proposed the vision-based lateral control framework in section IV. Section V and VI describe the MTL perception module and RL control module of the framework. The experiments are given in section VII. At last, we draw the conclusion and present future work in section VIII.
\begin{figure*}[!t]
	\centering
	\includegraphics[scale=0.5]{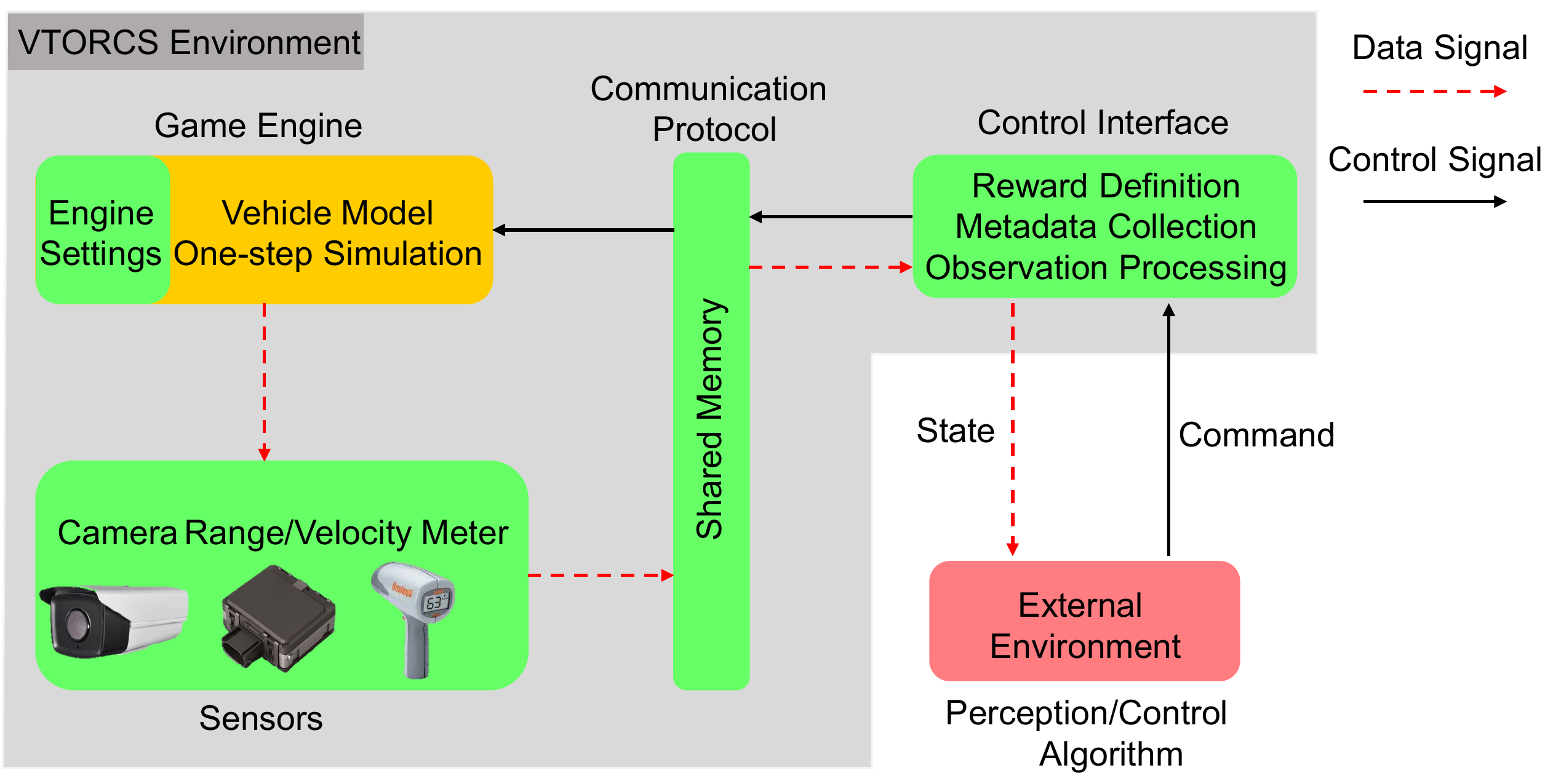}
	\caption{The VTORCS environment (light gray) includes four ingredients: the game engine, the sensors, the communication protocol, and the control interface.}
	\label{fig_vtorcs}
\end{figure*}

\section{The VTORCS Environment}
In this section, we give a detailed introduction to the VTORCS environment. As shown in Fig. \ref{fig_vtorcs}, the environment 
consists of four ingredients: a game engine block, a sensors block, a communication protocol block, and a control interface block. The green blocks in Fig. \ref{fig_vtorcs} are developed by us, and the yellow block is provided by the original TORCS. The game engine setting part is implemented based on the original TORCS in C++ language. Other components are developed in both C++ and Python. The environment provides Python interface for deep learning (DL) and RL algorithm (red block in Fig. \ref{fig_vtorcs}) implementation. The working flow includes two signal streams. In the data signal stream, the sensors firstly collect the measurement from the game engine and write all the data to shared memory via the communication protocol. The control interface then reads these data, assembles the observation, finally sends it to the external perception/control algorithm. In the control signal stream, the external algorithm computes and sends the control command to the control interface which will write the command to shared memory. Finally, the game engine reads the command from the communication protocol and finishes the one-step simulation. The details of these four ingredients are given below.

\subsection{Game Engine}
The core of VTORCS is the game engine where all simulation profiles are defined. It is responsible for vehicle model definition and simulation, graphics rendering, and control commands execution, etc. One of the most important features is the simulation frequency. One can define the desired control frequency through VTORCS engine settings, while the previous SCR TORCS fixes the control frequency to 5 Hz. A long control period will introduce delay and may cause the vehicle out of the track. We also provide a starter vehicle in the game engine, so one only needs to concentrate on the underlying algorithm implementation.

\subsection{Sensors}
In order to satisfy the needs of the perception and control algorithms, we provide various sensor supports. There are two categories of sensors in VTORCS: the high-dimensional camera sensor and the low-dimensional physical sensors. The former specializes in multi-view image capturing. However, different from the fixed resolution $ 64 \times 64 $ of SCR TORCS, our camera sensor can provide images with any desired size, for example, $ 320 \times 280 $ or $ 640 \times 480 $. The high-resolution image can provide more details which are crucial for the vision-based algorithm. In a low-resolution image, the distant region only occupies a few pixels and is often blurred, so the perception algorithm would be hard to extract the desired features. We also provide multiple physical sensor supports including range sensors, speed sensors, and angle sensors, etc. Moreover, one can customize the desired sensor easily through a sensor definition pipeline. All the sensor information is listed in Table \ref{tb_vtorcs_sensor}.

\begin{table}[!t]
	\renewcommand{\arraystretch}{1.3}
	\setlength{\tabcolsep}{1.4em}
	\caption{The VTORCS Sensors List.}
	\label{tb_vtorcs_sensor}
	\centering
	\begin{tabular}{l|c|c|c}
		\hline \hline
		\multicolumn{1}{c|}{Sensor} & Notation & Value & Unit \\
		\hline
		Range Meter & $ range $ &  [0, 200] & m \\
		\hline
		Angle Sensor & $angle$ & [$-\pi$, $\pi$] & rad \\
		\hline
		Speed Meter & $v$ & [0, 200] & km/h \\ 
		\hline
		Longitudinal Speed Meter & $v_x$ & [0, 200] & km/h  \\ 
		\hline
		Vertical Speed Meter & $v_y$ & [0, 50] & km/h  \\ 
		\hline 
		Engine Rotation & $rot$ & [0, 10000] & rpm \\ 
		\hline 
		Odometer & $len$ & [0, inf] & m  \\ 
		\hline
		Timer & $t_{lap}$ & [0, inf] & s \\
		\hline
		Wheel Spin Speed Meter & $ v_{wheel} $ & [0, inf] & rad/s \\
		\hline 
		Camera & $image$ & [0, 255] & - \\ 
		\hline \hline
	\end{tabular}
\end{table}

\subsection{Communication Protocol}
 For the purpose of interacting with the external perception/control module, a fast and dependable inter processing data transmitter is needed. The communication protocol, which is based on the shared memory is developed to meet these requirements. It is responsible for delivering raw sensor data to the control interface and transferring the control command to the game engine for execution. The employed shared memory is a special part of continuous memory space which can be simultaneously accessed by multiple programs. As a part of random access memory (RAM), the access speed of shared memory is extremely high. The related software commands for the shared memory manipulation include shm\_open, shmat, and shmget etc. To meet the dependable data transmission demands, a boolean flag is added to the end of every data message. Only when the flag shows that the data writing is completed, the control interface begins to prepare the observation and send it to the external module.

\subsection{Control Interface}
Since the raw sensor data is of different amplitudes and units, we design the control interface for data preprocessing and simulation statistics definition. The interface follows the popular OpenAI gym environment control pipeline which is logically explicit and specialized in RL algorithm implementation. In details, it includes four functions, i.e. state generation, reward definition, control command processing, and diagnostic information collection. The underlying control commands include steering, acceleration, brake, and gear change. The control interface block first parses sensor data from the shared memory when the data transmission process is finished. Then it normalizes the raw sensor signals and calculates the reward for the RL agent. Additionally, useful diagnostic information like vehicle running time, distance raced, etc. will also be collected. The control interface can be wrapped to a class so that the perception/control module can conveniently request the observation or send back the control command through it. Once a control signal is received, the interface will write it to the shared memory and finish the control pipeline. Note that one can also set a pre-programmed AI vehicle in the game engine and collect the desired dataset in this block.

\section{Problem Definition and Backgrounds}

\subsection{Vision-based Lateral Control Problem}
This paper aims to learn a lateral control policy with high-dimensional image stream as its input in the simulator VTORCS. The system is divided into two modules: the perception module and the control module. In the perception module, we try to train a neural network $ f_1(\cdot) $ to extract the track features:
\begin{equation}\label{abs_equ1}
\sigma_t = f_1(o_t; \mathbf{w}_1),
\end{equation}
where $ o_t $ is the observation i.e. driver-view image at time $ t $, $ \sigma_t $ is the track feature vector and $ \mathbf{w}_1 $ is the weight vector of perception neural network. Together with vehicle property $ \eta_t $, e.g. speed and engine rotation, etc., the control module maps the state $ s_t = [\sigma_t, \eta_t] $ to the action $ a_t $ via a neural network $ f_2(\cdot) $:
\begin{equation}\label{abs_equ2}
a_t = f_2(s_t;\mathbf{w}_2),
\end{equation}
where $ \mathbf{w}_2 $ is the weight vector for the control module. The goal is to optimize the weight vector $  \mathbf{w}=[ \mathbf{w}_1,  \mathbf{w}_2] $ so as to control the vehicle to drive along with the lane center. Additionally, we must restrict the computational complexity to satisfy the desired control frequency.

\subsection{Reinforcement Learning}
We solve the vision-based lateral control by employing RL methods. RL is a branch of machine learning and typically used to solve the sequence decision-making problem. In the RL settings, the problem is formulated as a Markov Decision Process (MDP) which is composed of a five-tuple $ (\mathcal{S}, \mathcal{A}, r(s_t,a_t), P(s_{t+1}|s_t, a_t), \gamma ) $. At time step $ t $, the agent selects the action $ a_t \in \mathcal{A} $ by following a policy $ \pi:\mathcal{S} \rightarrow \mathbb{R} $. After executing $ a_t $, the agent is transferred to the next state $ s_{t+1} $ with probability $ P(s_{t+1}|s_t,a_t) $. Additionally, a reward signal $ r(s_t, a_t) $ is received to describe whether the underlying action $ a_t $ is good for reaching the goal or not. For the purpose of brevity, rewrite $ r_t = r(s_t,a_t) $.  By repeating this process, the agent interacts with the environment and obtains a trajectory $ \tau = s_1,a_1,r_1,...,s_T,r_T $ at the terminal time step $ T $. The discounted cumulative reward from time step $ t $ can be formulated as $ R_t = \sum_{k=t}^{T}\gamma^{k-t}r_k $, where $ \gamma \in (0,1)$ is the discount rate that determines the importance of the future rewards. The goal is to learn an optimal policy $ \pi^* $ that maximizes the expected overall discounted reward
\begin{equation}\label{eq_rl_obj}
	J = \mathbb{E}_{s,a \sim \pi, r}[\sum_{k=1}^{T}\gamma^{k-1}r_k],
\end{equation}
\begin{equation}\label{rl_obj}
\pi^* = \argmax_{\pi} \mathbb{E}_{s,a\sim \pi, r}[R_1].
\end{equation}
Typically, two kinds of value functions are used to estimate the expected cumulative reward for a specific state
\begin{equation}\label{v_fn}
V^{\pi}(s) = \mathbb{E}_{\pi}[R_1|s_1=s],
\end{equation}
\begin{equation}\label{q_fn}
Q^{\pi}(s,a)=\mathbb{E}_{\pi}[R_1|s_1=s,a_1=a].
\end{equation}

The RL algorithms can be divided into two categories: the value-based algorithms and the policy-based algorithms. The most well-known value-based algorithm is temporal difference (TD) learning which is a set of algorithms including Q-learning\cite{Watkins1989Learning}, SARSA\cite{Rummery94on-lineq-learning}, and TD($ \lambda $)\cite{Sutton1988Learning}. In these algorithms, the agent estimates a value function (\ref{v_fn}) or (\ref{q_fn}), and the control policy is generated based on the value function. In the policy-based (a.k.a. policy gradient) algorithms, the optimal policy is directly approximated by optimizing the objective function which typically is the discounted cumulative reward (\ref{eq_rl_obj}). The popular policy-based algorithms include REINFORCE\cite{Williams1992Simple} and DPG\cite{Silver2014Deterministic}.

\subsection{Multi-task Learning}
Deep neural networks, such as CNNs, usually need tremendous samples and time to converge. Training a CNN from scratch is prone to overfit when the dataset is small and noisy. MTL\cite{Caruana1997Multitask} is a category of popular methods whose principal purpose is to reduce overfitting and improve model generalization performance. In the MTL framework, $N$ related tasks $\{\mathcal{T}_i\}_{i=1}^N$ are solved jointly. While exploiting similarities and differences among tasks, MTL tends to improve the learning of task $\mathcal{T}_i$ by utilizing the knowledge of other tasks. For the task $\mathcal{T}_i$, a dataset $ \mathcal{D}_i = \{(o_{i,j}, y_{i,j})\}_{j=1}^{N_i} $ is given where  $o_{i,j}$ is the $j$th observation, $y_{i,j}$ is the corresponding label, and $ N_i $ is the number of samples. For example, for the distance to lane marking prediction task, $o_{i,j}$ is the driver-view image and $y_{i,j}$ are the distance to lane markings. Assume the loss function for task $\mathcal{T}_i$ is $\sum_{j=1}^{N_i}\mathcal{L}_i(y_{i,j},f(o_{i,j}; \mathbf{w}_1^i))$ where $ \mathcal{L}_i(\cdot,\cdot) $ is a classification or regression loss function and $\mathbf{w}_1^i$ are the weights for the task $\mathcal{T}_i$. Then the MTL loss can be formulated as:
\begin{equation}\label{abs_mtl}
\mathcal{L}_{mtl} = \sum_{i=1}^{N}\alpha_i \mathcal{L}_i(y_{i,j}, f(o_{i,j}; \mathbf{w}_1^i)) + \varPhi(\mathbf{w}_1^i).
\end{equation}
Here $ \alpha_i $ is a hyper-parameter that adjusts loss proportion, and $ \varPhi(\cdot) $ is a  $l_1$ or $l_2$ norm regulation.

The MTL methods mainly focus on the representation of shared features and the way to share it. One popular way to represent the shared knowledge is by employing neural network. The knowledge is fused into the hidden layer features and shared among all tasks. This kind of approach is usually based on the deep neural networks and solves multiple tasks end-to-end. Another way to represent the knowledge is the feature learning approach for shallow models where the common features are learned under the regularization framework. The loss function is similar to (\ref{abs_mtl}) excepting the regulation function is replaced by a $l_{p,q}$ norm denoted by $||\mathbf{W}||_{p,q}$. For a little abuse of notations, here $\mathbf{W}$ is the weight matrix containing weights for all tasks in the shallow model where each column $\mathbf{w}^i$ contains the weights for task $\mathcal{T}_i$ and each row $\mathbf{w}_j$ contains the weights for the $j$th feature across all tasks. Then the $l_{p,q}$ norm is defined as $||\mathbf{W}||_{p,q} = ||(||\mathbf{w}_1||_p,...,||\mathbf{w}_N||_p)||_q$ where $||\cdot||_p$ is the $l_p$ norm of a vector. The $l_{p,q}$ norm regulation is based on the group-sparsity\cite{Argyriou2006nips} which assumes only a small set of features are shared across all tasks. Therefore, the weight matrix $\mathbf{W}$ tends to be row-sparse so as to select essential features. The popular choices of $l_{p,q}$ norm include $l_{2,1}$ norm in \cite{Seunghak2010nips} and $l_{\infty,1}$ in \cite{Liu2009icml}.

There are many ways of sharing features among tasks. The most common approach\cite{long2017nips} is to share the low-level features and separate the high-level task-specific features. Based on this feature sharing manner, \cite{Kendall2017End} utilizes the uncertainty of each task to adjust the loss proportion $\alpha_i$ in (\ref{abs_mtl}). In \cite{Lu2017Fully}, a fully-adaptive feature sharing network is proposed which dynamically broadens the network branches from the output to input layers by grouping similar tasks. In the above methods, only one neural network is used to tackle all tasks. Differently, a cross-stitch unit is proposed in \cite{MisraCrossMTL16} to linearly combine the intermediate features between multiple neural networks.

\begin{figure}[!t]
	\centering
	\includegraphics[scale=0.55]{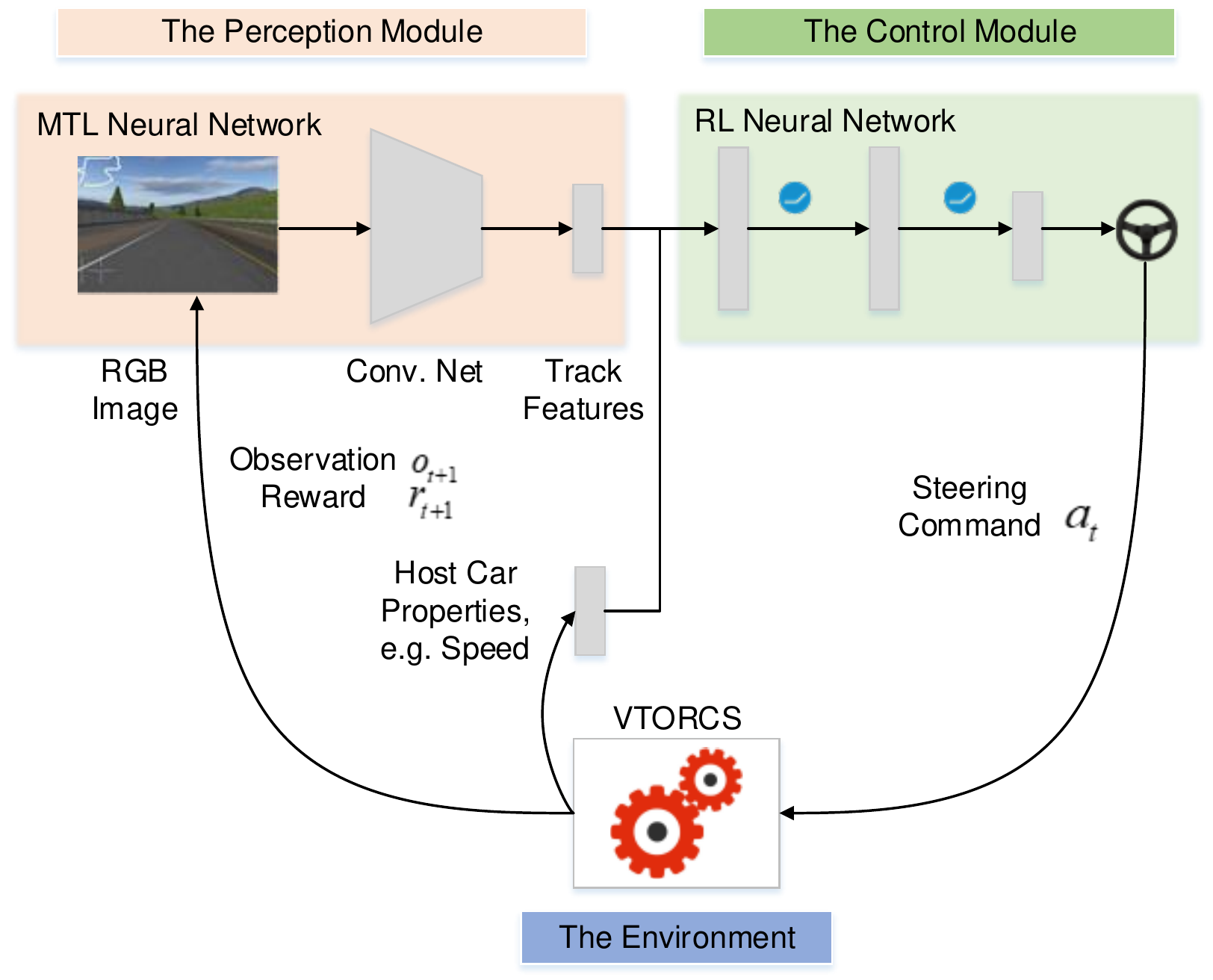}
	\caption{The vision-based lateral control framework which includes: the perception module, the control module, and the VTORCS environment.}
	\label{fig_sys_framework}
\end{figure}

\section{Vision-based  Lateral Control Framework}
In this section, we present an overview of the vision-based lateral control framework. As shown in Fig. \ref{fig_sys_framework}, it consists of three parts: the MTL perception module, the RL control module, and the VTORCS environment. 

We treat the problem as a discrete-time perception and control problem. At every time step $ t $, the VTORCS environment sends an RGB image to the MTL perception network. After perception, the MTL network transmits the track features to the RL control module which outputs the desired steering command and feeds it back to the VTORCS. Finally, the game engine executes the control command and finishes the one-step simulation. In details, the MTL perception network works in a global coordinate and takes image $ o_t $ as its input. According to (\ref{abs_equ1}), it maps high-dimensional variable to a low-dimensional track feature vector $ \sigma_t $ by solving multiple tasks jointly. Note that the track features are defined in track coordinate in which the origin is at the track horizontal center, the X axis is along the track heading direction, and the Y axis is vertical to the track heading. We combine the feature vector $ \sigma_t $ and vehicle property vector $ \eta_t:= [v_{x,t}, v_{y,t}] $  together to form the RL state variable $ s_t $. The notations $v_{x,t}$ and $v_{y,t}$ represent the vehicle longitudinal and lateral speed at time $t$, respectively. In order to integrate multiple values with different units, we normalize the state $s_t$ to $[0,1]$ by dividing each element of $s_t$ by the corresponding maximum value, e.g. half track width for distance to lane center, $\pi$ for yaw angle, and $75$ km/h for vehicle speed. For the purpose of being robust to perception noise and easy to transfer, RL control network works in the track coordinate. It maps state variable $ s_t $ to the control action $ a_t $ by following (\ref{abs_equ2}). In practice, we perturb $ s_t $ by a Gaussian noise $ \epsilon \sim \mathcal{N}(0, \Sigma_n) $ to introduce an additional disturbance to sensors. Finally, the action $ a_t $ is delivered to the game engine via VTORCS communication protocol, which also returns the next observation $ o_{t+1} $ and reward $ r_{t+1} $.

\section{Multi-task Learning Traffic Scene Perception}
In this section, we first analyze and select the proper tasks for MTL. Then we define the corresponding loss function and introduce the network architecture. 

\subsection{Choice of Related Tasks}
The goal of this module is to predict the accurate and robust track features that are essential for control. To this end, we first need to select the desired features. In a scene of VTORCS (as shown in Fig. \ref{fig_track_scene}), there are many related tasks, which include the distance to lane markings, the heading angle differences, the track heading direction, the number of lanes, and the track curvature, etc. Since the control algorithm works in the track coordinate, we must locate the vehicle in that coordinate. Here we select the first three tasks for the purpose of localization. By combining the distance to lane markings $ {d_1, d_2, \cdots, d_N} $ and the angle between track and vehicle heading direction $ \theta $, we can obtain the vehicle location and pose. The third task which classifies the track to the type of left/right curve and straight road is able to describe current track direction at a high level. This feature can give a sense that which kind of steering command should take in the next few time steps. 

\subsection{MTL Model} 
The MTL model solves the above three related tasks at the same time. Assume we have a set of observations $ \{o_i\}_{i=1}^N $ and the corresponding labels $ \{y_i^1, y_i^2, y_i^3\}_{i=1}^N $. $ y_i^1, y_i^2, y_i^3 $ represent the ground-truth for distance to lane markings, heading angle difference, and track heading type, where the superscript represents the task number and the subscript $ i $ represents the sample index. Since the first two are regression tasks and the third one is a classification task, we employ the least square loss for regression tasks and the cross-entropy loss for the classification task. In details, the task 1 takes the image and predicts five normalized distances to the lane markings, which include the distance to the left/right lane markings when the vehicle runs in the lane, and the distance to the left/right markings of the neighbor lane and the distance to the current lane marking when the vehicle runs on the lane marking. The distance loss is defined as:
\begin{equation}\label{eq_dist_loss}
\mathcal{L}_1 = \frac{1}{N_1}\sum_{i=1}^{N_1}(y_{i}^{1} - f_1(o_i;\mathbf{w}_{1}^{1}))^2,
\end{equation}
where $ N_1 $ is the number of samples and $ f_1(o_i;\mathbf{w}_{1}^{1}) $ is the network output for the observation $ o_i $. Recall from (\ref{abs_equ1}), $ \mathbf{w}_{1} $ represents the weight of all perception neural networks, so $ \mathbf{w}_{1}^{1} $ stands for the weight of task 1, i.e. the distance to markings regression task. 

The task 2 predicts the normalized angle difference between the vehicle and the track heading. Thus, the output dimension of this task is 1. The loss is defined as:
\begin{equation}\label{eq_angle_loss}
\mathcal{L}_2 = \frac{1}{N_2}\sum_{i=1}^{N_2}(y_{i}^{2} - f_1(o_i;\mathbf{w}_{1}^{2}))^2.
\end{equation}

The task 3 classifies the track heading direction into three categories: the left or right turn and the straight. The loss is defined as:
\begin{equation}\label{eq_trk_type}
\mathcal{L}_3 = \frac{1}{N_3}\sum_{i=1}^{N_3}-\log(f_1(o_i;\mathbf{w}_{1}^{3})_{y_{i}^3}),
\end{equation}
where $ f_1(o_i;\mathbf{w}_{1}^{3})_{y_{i}^3} $ is the $ y_i^3 $th element of the task 3 network softmax layer's output. By combining the above three losses together, the MTL neural network is solving the following optimization problem
\begin{equation}\label{eq_mtl_obj}
\argmin_{\mathbf{w}_1} \sum_{t=1}^{3}\alpha_t \mathcal{L}_t + \varPhi(\mathbf{w}_1).
\end{equation}
Here $ \mathbf{w}_1=[\mathbf{w}_1^1, \mathbf{w}_1^2, \mathbf{w}_1^3] $ is the weight of all the MTL perception network, and $ \alpha_t $ is the coefficient to weigh the loss of task $ t $. The last term $ \varPhi(\mathbf{w}_1) = \lVert \mathbf{w}_1 \rVert_2^2 $ is the $l_2$-norm regulation function which is typically used to reduce overfitting and improve model generalization performance.

\subsection{Network Structure}
\begin{figure}[!t]
	\centering
	\includegraphics[scale=0.55]{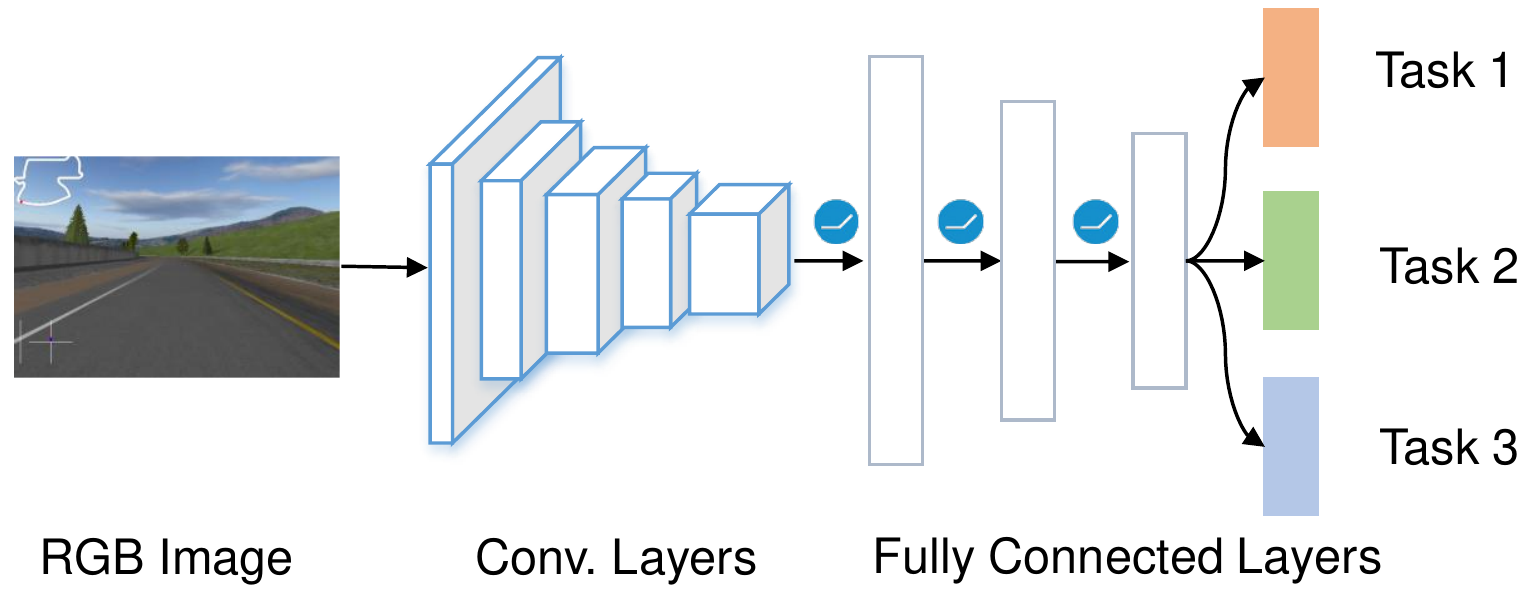}
	\caption{Multi-task learning neural network architecture. MTL network consists of 5 convolutional layers (Conv. layers) and 4 fully connected layers.}
	\label{fig_mtl_nn}
\end{figure}
The MTL network architecture is shown in Fig. \ref{fig_mtl_nn}, which includes two parts. The network takes the $ (280, 210, 3) $ dimensional RGB image as its input and solves the three perception tasks at the same time. The low-level convolutional layers and fully connected layers are shared across three tasks to extract the general features. These features are not sensitive to a specific task and essential for reducing the overall combined loss. This part includes five convolutional layers whose configurations are as follows: Conv(96, 11, 4) -- MaxPool(3, 2) -- Conv(256, 5, 2) -- MaxPool(3, 2) -- Conv(384, 3, 2) -- Conv(384, 3, 2) -- Conv(256, 3, 2) -- MaxPool(3, 2). The notation Conv($ n $, $ k $, $ s $) represents a convolutional layer that has $ n $ filters with size $ k \times k $ and stride $ s $. The MaxPool($ k $, $ s $) expresses the max-pooling operation with kernel size $ k \times k $ and stride $ s $. On the top of convolutional layers, there are three shared fully connected layers, which contain 4096, 1024, and 256 hidden units, respectively. Then three fully connected layers for specific tasks are built on the shared layers. The output layers include 5, 1, and 3 neurons according to the task specification introduced in the sub-section B. For all hidden layers, we use $ ReLU $ activation function $ \sigma(x) = \text{max}(0, x) $. For the classification task 3, the last layer activation function is softmax
\begin{equation}\label{fn_softmax}
\sigma(x_j) = \frac{\exp(x_j)}{\sum_{k=1}^{K}\exp (x_k)}.
\end{equation}
The last layer of the two regression tasks take the linear activation function. To combat overfitting, we employ the Dropout layer\cite{Srivastava2014Dropout} after all fully connected layers.

\section{Reinforcement Learning Lateral Control}
Our goal is to control the vehicle running along the lane center based on the image. After obtaining the perceived track features, we define our formulation for the lateral control problem in this section. First, we introduce the RL control algorithm, and then give the learning setup in details.

\subsection{Deterministic Policy Control Algorithm}
In the lateral control problem, the agent makes a decision, issues the action in VTORCS, and receives the next observation and evaluative reward. It repeats this process at every time step. We utilize RL to tackle this sequence decision-making problem. In VTORCS, the steering command is a continuous value in $ [-1,1] $. Therefore, the lateral control is a continuous action control problem. We employ recently developed continuous action control RL algorithm: deterministic policy gradient (DPG)\cite{Silver2014Deterministic}. Different from the stochastic policy, the deterministic policy $ \mu(s_t) $ outputs a scalar action and executes it with probability 1, while the stochastic policy $ \pi(s_t,a_t) $ outputs the stochastic probability for every action.

The DPG aims to directly search an optimal policy $\mu^*(s_t)$ in the policy space that maximizes the objective $J$ defined in (\ref{eq_rl_obj}). Here we utilize the actor-critic approach to approximate the optimal policy in the neural dynamic programming framework\cite{si2001online}. The actor $\mu(s_t;\mathbf{w}_2^{\mu})$ where $\mathbf{w}_2^{\mu}$ are the network weights approximates the optimal policy. The critic $Q^{\mu}(s_t, a_t; \mathbf{w}_2^Q)$ where $\mathbf{w}_2^Q$ are the network weights approximates the optimal action-value function. Since the actor aims to maximize the objective $J$, the network weights $\mathbf{w}_2^{\mu}$ are updated by using the stochastic gradient ascent algorithm
\begin{equation}\label{eq_actor_update}
\mathbf{w}_2^{\mu} \leftarrow \mathbf{w}_2^{\mu} + \alpha_{\mu} \nabla_{\mathbf{w}_2^{\mu}}J.
\end{equation}
where $\alpha_{\mu}$ is the learning rate of actor, and $\nabla_{\mathbf{w}_2^{\mu}}J$ are the gradients of the objective $J$ with respect to actor weights  $\mathbf{w}_2^{\mu}$. The gradients $\nabla_{\mathbf{w}_2^{\mu}}J$ are obtained by following the deterministic policy gradient theorem\cite{Silver2014Deterministic}
\begin{equation}\label{eq_actor_gradient}
\nabla_{\mathbf{w}_2^{\mu}}J = \mathbb{E}_{s_t}[\nabla_a Q(s_t,a;\mathbf{w}_2^Q)|_{a=\mu(s_t)} \nabla_{\mathbf{w}_2^{\mu}}\mu(s_t;\mathbf{w}_2^{\mu})].
\end{equation}

In the critic network, we approximate the action-value function iteratively by following the Bellman equation
\begin{equation}\label{dpg_Q}
Q^{\mu}(s_t, a_t; \mathbf{w}_2^Q) = \mathbb{E}_{s_{t+1}}[r_t + \gamma Q^{\mu}(s_{t+1}, \mu(s_{t+1}; \mathbf{w}_2^{\mu});  \mathbf{w}_2^Q)].
\end{equation}
It is generally believed that using the nonlinear neural network to represent the action-value function $Q(s_t, a_t)$ is unstable or even to diverge. In order to tackle the instabilities, we employ two techniques as in \cite{mnih2015human-level}, i.e. experience replay\cite{lin1993reinforcement} and target networks. The experience replay holds a finite buffer $D$ containing experience $(s_t,a_t,r_t,s_{t+1})$ at each time-step. At training phase, a mini-batch experiences are uniformly sampled from the buffer to update the critic network weights. The target networks are the same as the actor and critic networks excepting the target network weights $\mathbf{w}_2^{\mu-}, \mathbf{w}_2^{Q-}$ are slowly and "softly" copied from the actor and critic weights, i.e. $\mathbf{w}_2^{\mu-} \leftarrow \tau \mathbf{w}_2^{\mu} + (1-\tau)\mathbf{w}_2^{\mu-}$, $\mathbf{w}_2^{Q-} \leftarrow \tau \mathbf{w}_2^{Q} + (1-\tau)\mathbf{w}_2^{Q-}$, and $\tau \ll 1$. This slowly updating property keeps the target of critic more stable and formulates a stable optimization problem. Since the critic approximates the action-value function by using the Bellman equation, this can be achieved by minimizing the mean square loss
\begin{equation}\label{eq_critic_loss}
\mathcal{L}(\mathbf{w}_2^Q) = \mathbb{E}_{s_t, a_t, r_t}[(y_t - Q^{\mu}(s_t, a_t; \mathbf{w}_2^Q))^2]
\end{equation}
where the target $ y_t $ is
\begin{equation}\label{eq_critic_target}
y_t = r_t + \gamma Q^{\mu}(s_{t+1}, \mu(s_{t+1};\mathbf{w}_2^{\mu-}) ; \mathbf{w}_2^{Q-}).
\end{equation}
By computing the gradients of the loss function (\ref{eq_critic_loss}) with respect to the weights $\mathbf{w}_2^{Q}$, the critic network is updated by following the stochastic gradient descent algorithm
\begin{equation}\label{eq_critic_update}
\mathbf{w}_2^{Q} \leftarrow \mathbf{w}_2^{Q} - \alpha_{Q} \nabla_{\mathbf{w}_2^{Q}} \mathcal{L}(\mathbf{w}_2^Q)
\end{equation}
where $\alpha_{Q} $ is the learning rate of critic, and the gradients are obtained on the mini-batch with size $m$
\begin{equation}\label{eq_critic_gradient}
\nabla_{\mathbf{w}_2^{Q}} \mathcal{L}(\mathbf{w}_2^Q) = \dfrac{1}{m} \sum_{i=1}^{m} (y_i - Q^{\mu}(s_i, a_i; \mathbf{w}_2^Q)) \nabla_{\mathbf{w}_2^{Q}}Q^{\mu}(s_i, a_i; \mathbf{w}_2^Q).
\end{equation}

\begin{figure}[!t]
	\centering
	\includegraphics[scale=0.65]{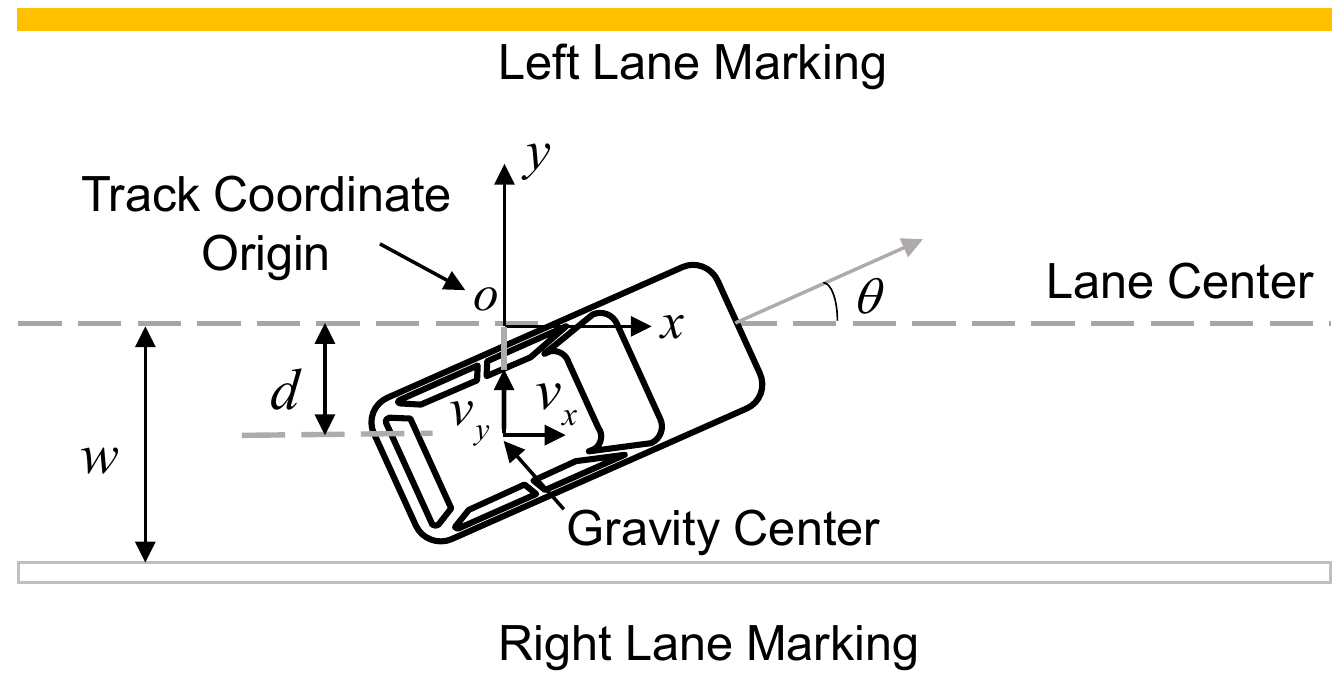}
	\caption{Lateral control reward design. The agent to minimize the yaw angle $ \theta $ so as to keep the vehicle heading along track heading direction.}
	\label{fig_reward}
\end{figure}

\subsection{Algorithm Stability Property}
To the best of our knowledge, the stability guarantee of RL algorithms is still a challenge when representing the action-value function with the nonlinear neural network. The main reasons are i) the correlated sample distribution which breaks the independently and identically distributed (IID) assumption required by most optimization algorithms; and ii) the instable optimization target value which changes quickly by using the same weights as the critic. To solve these two problems, we employ two techniques, i.e. experience replay and target network as in\cite{mnih2015human-level}. At each training step, a mini-batch is uniformly sampled from the replay buffer, and it will randomize the sampled experiences in order to satisfy the IID assumption on which most optimization algorithms are based. Additionally, by employing the target network and constraining the change of the target of critic's loss function, the target value is more stable and the optimization of the critic is more like a supervised learning problem. By employing these two techniques in practice,the training process is stable and the agent is able to learn a good lateral control policy.

\subsection{System Parameters}
Here we introduce the key ingredients of the later control RL algorithm, which contain the state representation, the action definition, and the reward design.

\subsubsection{State}
After forward propagation of the MTL neural network, we obtain the track feature vector $ \sigma_t $ which includes distance to lane center $d_t$, heading angle difference $\theta_t$, and track heading direction. These features are sufficient to determine vehicle's location and pose in a static scene. Additionally, in some cases, we also need to augment the above features with the vehicle property $ \eta_t $. For example, the speed at a curve is significant for keeping the vehicle running at the lane center. We need to determine proper steering angles according to different speeds. Therefore, the augmented features consist of vehicle speed along track direction $ v_{x,t} $ and vertical to track direction $ v_{y,t} $, i.e. $ \eta_t = [v_{x,t}, v_{y, t}] $. In order to model the disturbance to sensors and perception errors, we add a Gaussian noise $ \mathcal{N}(0, 0.05^2) $ to the normalized state $ s_t = [\sigma_t, \eta_t] $.

\subsubsection{Action}
In the lateral control problem, the action $ a_t $ is continuous normalized steering angle in $ [-1, 1] $ by dividing the steering ratio where the negative value is for turning right and the positive value is for turning left. In order to balance exploitation and exploration, we take $ \epsilon $-greedy policy in (\ref{eq_eps_greedy}) during the training process. The deterministic behavior policy $ \mu(s_t;\mathbf{w}_2^{\mu}) $ is perturbed by a Gaussian noise with probability $ \epsilon $:
\begin{equation}\label{eq_eps_greedy}
a_t = \begin{cases}
	\mu(s_t;\mathbf{w}_2^{\mu}), & \text{if $ p > \epsilon $} \\
	\mu(s_t;\mathbf{w}_2^{\mu}) + \beta \mathcal{N}(0, 0.05^2), & \text{otherwise}.
\end{cases}
\end{equation}
The coefficient $ \beta $ is used to adjust perturbation level. During the training phase, the initial value is $ \epsilon_{init} = 1.0 $ . As in (\ref{eps_decay}), the explore rate $ \epsilon $ decays linearly with the minimum $ 0.1 $ which is used to keep a low-level exploration. The probability $ \epsilon $ will decrease to $ 0.1 $ when training step $ t $ reaches the maximum exploration steps $ T_{\epsilon} = 4\times 10^5 $.
\begin{equation}\label{eps_decay}
\epsilon = \max(0.1, \epsilon_{\text{init}} - 0.9\frac{t}{T_{\epsilon}}).
\end{equation}

\subsubsection{Reward}
As a key element of RL framework, the reward signal drives the agent to reach the goal by rewarding good actions and penalizing poor actions. In the lateral control task, the goal is to control the vehicle run in the track center. We design reward function in track coordinate. Additionally, to provide a persistent learning signal, the reward is provided at every step instead of at the end of an episode. As shown in Fig. \ref{fig_reward}, reward is a function of the distance to lane center $ d $ and the angle between vehicle heading direction and lane center $ \theta $:
\begin{equation}\label{eq_reward}
	r=\begin{cases}
	\cos(\theta) - \lambda\sin(\lvert \theta \rvert) - d/w, & \text{if $ \lvert \theta \rvert < \frac{\pi}{2} $},\\
	-2, & \text{otherwise},
	\end{cases}
\end{equation}
where $ w $ is the half width of lane, and $ \lambda $ is a coefficient that adjusts the horizontal and vertical influence of $ \theta $. Note $ \theta $ is an acute angle with sign, i.e. $ \theta \in [-\frac{\pi}{2}, \frac{\pi}{2}] $, and $d \in [0, \infty)$ is the magnitude of the distance. The positive value is for a clockwise angle, and the negative value is for an anticlockwise angle. When the vehicle runs out of track or runs backward, we terminate the episode and penalize the action with a high penalty $ -2 $.

\begin{figure}[!t]
	\centering
	\includegraphics[scale=0.6]{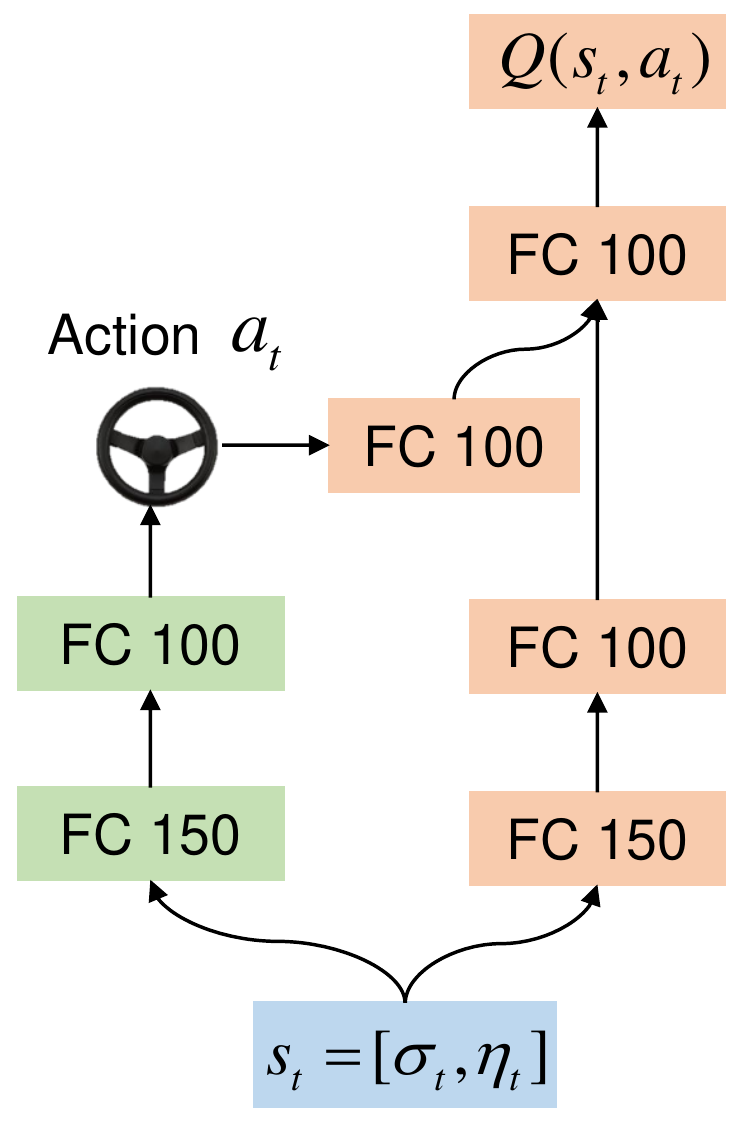}
	\caption{RL lateral control network architecture.} 
	\label{fig_rl_nn}
\end{figure}

\subsection{Network Structure}
The RL control network architecture is shown in Fig. \ref{fig_rl_nn} where $ FC \ num $ stands for the fully connected layer with $ num $ hidden units. The architecture consists of two networks, the actor network (in green) for selecting an action and the critic network (in red) for evaluating the action in the underlying state. The network input is $ s_t = [\sigma_t, \eta_t] $ that includes MTL perception network outputs $ \sigma_t $ and vehicle property $ \eta_t $. The action is fed into critic network until the last layer.

\section{Experiments}
In this section, we validate the vision-based lateral control algorithms in two aspects, which include MTL traffic scene perception and RL control. We evaluate the MTL network in various VTORCS tracks, and also compare its performance with the single task learning. One majority advantage of MTL is its robust perception accuracy in different scenes like in sunny or shadow conditions. The RL controller is trained and compared with other methods in various tracks.

\subsection{Dataset Collection}
\begin{figure}[!t]
	\centering
	\includegraphics[scale=0.45]{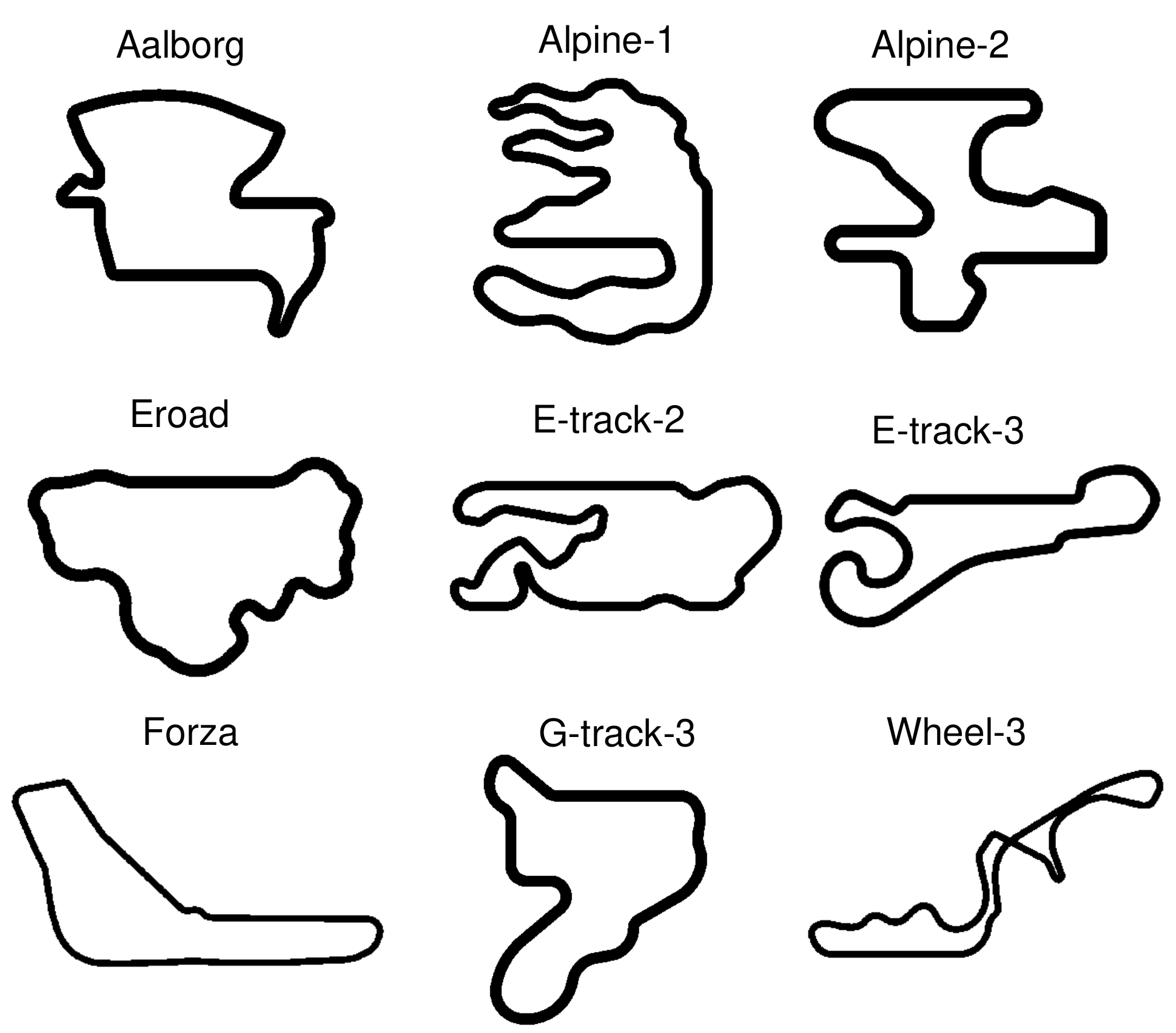}
	\caption{Various tracks for data collection.}
	\label{fig_data_collector}
\end{figure}

We implement all experiments by using Tensorflow on a computer with an Intel Xeon E5-2620 CPU and Nvidia Titan Xp GPU. To train the MTL neural network, a preprogrammed AI vehicle is used to collect the driver-view images and corresponding labels in a rich set of tracks. As shown in Fig. \ref{fig_data_collector}, we totally use nine tracks including alpine-1, alpine-2, g-track-3, and e-track-3, etc. to accomplish data collection task. To increase diversity of the dataset, all the tracks are set with one, two, and three lanes in each data collection round. For a specific track, the predefined trajectories are set differently for the AI vehicle in different data collection rounds. We also set different amount of traffic vehicles to add dataset diversity. Finally, we collect about 120k images in total. Then we pick samples of 15 rounds (about 30k images) as the testing dataset and blend samples of other rounds as the training dataset (about 90k images). Note this dataset blend procedure means the testing samples are collected in the different rounds compared with the training samples. This will guarantee no images between the testing dataset and the training dataset are very similar, and keep the testing dataset valid.

\subsection{Multi-task Learning Results}
\begin{figure}[!t]
	\centering
	\includegraphics[scale=0.45]{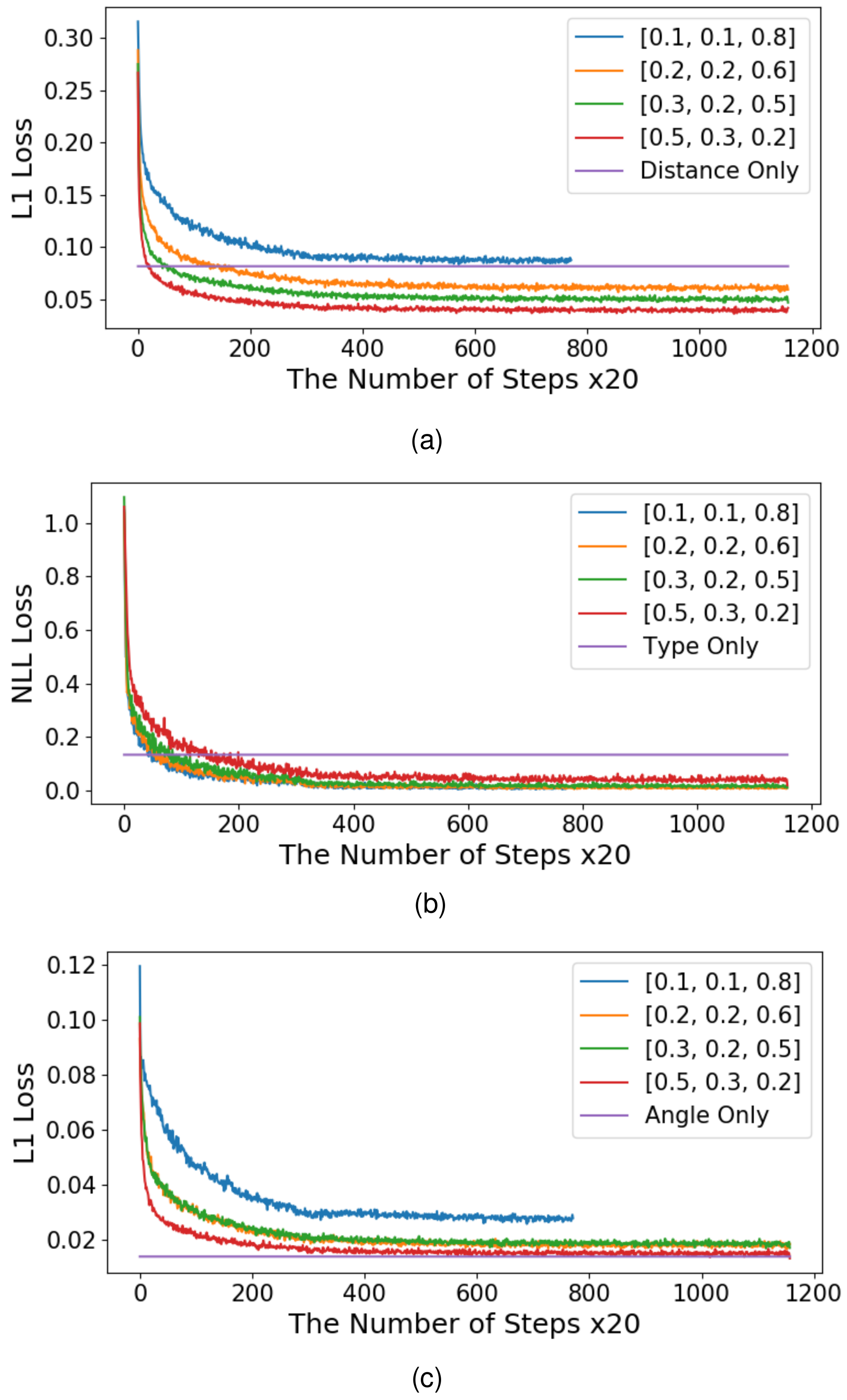}
	\caption{MTL training losses for three tasks. The figure (a), (b) and (c) show training losses for distance to lane markings, angle difference and track heading type classification, respectively. The horizontal purple line is the final loss for the single task learning and others are MTL losses with different coefficient settings.} 
	\label{fig_mtl_loss}
\end{figure}

In MTL experiments, the network is trained using stochastic gradient descent optimizer with momentum $ m=0.9 $. The batch size is set to 32. The initial learning rate is chosen as 0.001 and decays by a factor of 0.9 every 10k steps. We train the network for 150k iterations in each experiment, which takes 7.5 hours to finish training. Fig. \ref{fig_mtl_loss} shows the training loss curves with different loss coefficients. The thin purple horizontal line is the final training loss of the single task learning at the same training iteration as MTL, and others correspond to losses with different coefficient $ \alpha $ configurations.

We can see in all three tasks, MTL can achieve a lower loss. Especially in Fig. \ref{fig_mtl_loss} (a) and (c), MTL outperforms single task learning with a large margin. This interprets that MTL network is capable of capturing more general features across tasks that are essential for solving the underlying learning problem. Another interesting fact is that MTL network can reach a lower training loss for a specific task if we increase its proportion $ \alpha $.

\begin{table}[!t]
	\renewcommand{\arraystretch}{1.3}
	\caption{Multi-task Learning Test Loss with Various Settings}
	\label{tb_mtl_test_loss}
	\centering
	\begin{tabular}{c|c|c|c|c|c|c}
		\hline \hline
		\, & \multicolumn{3}{c|}{Task Loss Coefficient $ \alpha $} & Distance & Angle & Type \\ 
		\cline{2-7}
		\,       & Distance &   Angle  &   Type  & Loss $ l_1 $ & Loss $ l_2 $ & Accuracy \\
		\hline
		Distance &	   1	&	  0	   &	0	 &	 0.06771    &  	 -         & - \\
		Angle	 &     0    &     1    &    0    &   -          &    0.01304   & - \\
		Type     &	   0    &     0    &    1    &   -          &    -         & \textbf{99.24\%}\\
		\hline
		MTL 1 &  0.1  &    0.1   &   0.2   &   0.03841    &    0.01326   &  99.14\%  \\
		MTL 2 &  0.2  &    0.2   &   0.6   &   0.02655    &    0.01233   &  99.15\%  \\
		MTL 3 &  0.3  &    0.2   &   0.5   &   0.02253    &\textbf{0.0115}&  98.78\% \\
		MTL 4 &  0.5  &    0.3   &   0.2   & \textbf{0.01739}& 0.01838   &  98.03\%  \\
		\hline \hline
	\end{tabular}
\end{table}

Table \ref{tb_mtl_test_loss} shows both single task learning and MTL losses on the testing dataset. The upper half of the table holds the losses for single task learning while the lower half is for MTL. The best performance for a specific task is in boldface. All the losses are obtained with the same training iteration. For the task 1, i.e. the distance to lane markings, MTL achieves the best loss 0.01739 which is more than three times lower than learning this task alone. For the angle task, MTL can also obtain better performance than training task alone. For the classification task, MTL accuracy is slightly lower than the single task learning, but both of them are above 99\%. Focus on the MTL loss for a particular task, for example, the angle task, the loss is like an anti-bell shape. It drops at the beginning when the coefficient $\alpha$ increases, and increases when $ \alpha $ exceeds a particular value.

\begin{figure}[!t]
	\centering
	\includegraphics[scale=0.55]{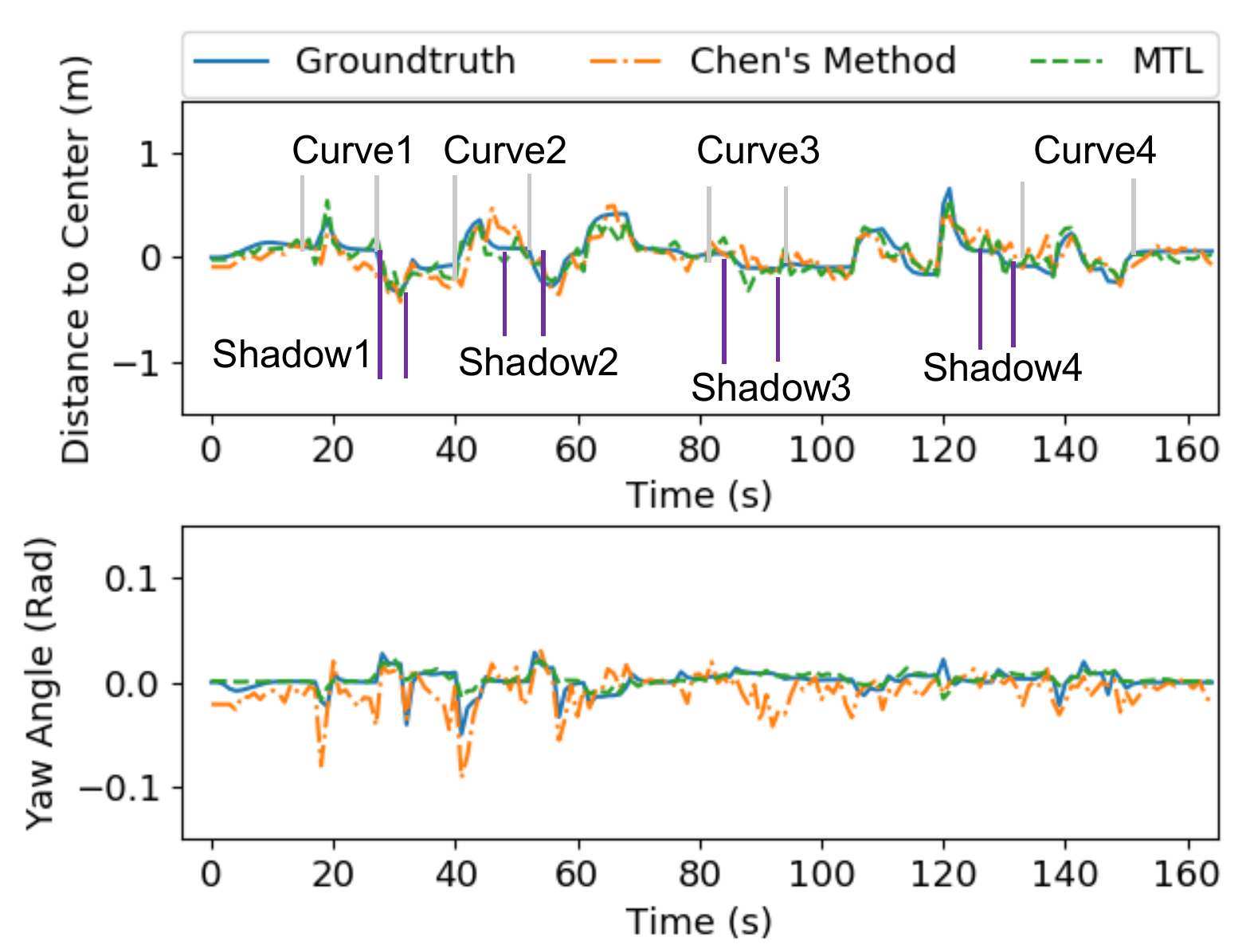}
	\caption{The perception performance comparison between the proposed MTL network and Chen's single task learning network.}
	\label{fig_mtl_shadow}
\end{figure}

The motivation of introducing MTL is to obtain better perception performance. Here we compare the performance of the proposed MTL algorithm and the Chen's DeepDriving\cite{chen2015deepdriving} perception network which is a single task learning network.  For fair comparison, an AI vehicle drives along the track and collect images and labels. Thus, the input is the same for two comparison methods. The experiment is implemented in track g-track-3 and the result is shown in Fig. \ref{fig_mtl_shadow}. The above figure illustrates the distance to lane center in meters, and the bottom figure illustrates the yaw angle in radian. By introducing the track type classification task, MTL achieves better perception performance. The reason is that the crucial features in classification task are also essential for other two tasks. For example, in the straight road, the distance and angle vary in a small range causing the accurate regression a hard task. However, the MTL network must perceive the small shifting features of road to make a right classification. Due to the shared network architecture in MTL, these features will further facilitate two regression tasks and improve the prediction accuracy. Especially in four sharp curves marked in Fig. \ref{fig_mtl_shadow}, the proposed MTL method (in green) fits the ground-truth more accurately than Chen's method (in orange). This is crucial for vision-based control because poor perception will cause controller predicting jittering steering command. Under some circumstances, such as at a shape curve, the vehicle may run out of the track. Additionally, the four-time intervals of heavy shadow marked in Fig. \ref{fig_mtl_shadow} show that the MTL perception module is able to work robustly in the poor lighting condition. The mean distance prediction error is 0.071 m and the mean yaw angle prediction error is 0.003 rad in these intervals. To validate whether the MTL perception network can transfer to the unknown track, we test it on an unknown track dirt-3, the result shows that the mean distance to middle prediction error is 0.107m and the mean yaw angle error is 0.004 rad. This performance is comparable to the results on the known track.

\subsection{Reinforcement Learning Lateral Control Results}
In the RL lateral control experiments, we train a deterministic policy to realize continuous action control. In all experiments, the discount rates $ \gamma=0.99 $, and the optimizer is Adam\cite{Kingma2015adam} optimizer. The learning rate for the actor network and the critic network are selected as $ 1e^{-3} $ and $ 1e^{-4} $, respectively. An episode is terminated if the vehicle runs out of track, runs backward, or the time step reaches its maximum $6\textrm{,}500$. The simulation frequency is 20 Hz. In all experiments, we fix the gear to 1 and throttle to 0.2 which will constrain the vehicle speed in [60, 75]km/h. The required training time varies among tracks because of the different difficulty levels. For the easy track like forza, it takes 18.3 minutes to learn the policy. The difficult track alpine-2 costs the most time to train, which is 45.8 minutes. However, thanks to the good generalization performance, the RL controller trained in the track forza can control the vehicle in the lane in alpine-2.

\begin{figure}[!t]
	\centering
	\includegraphics[scale=0.56]{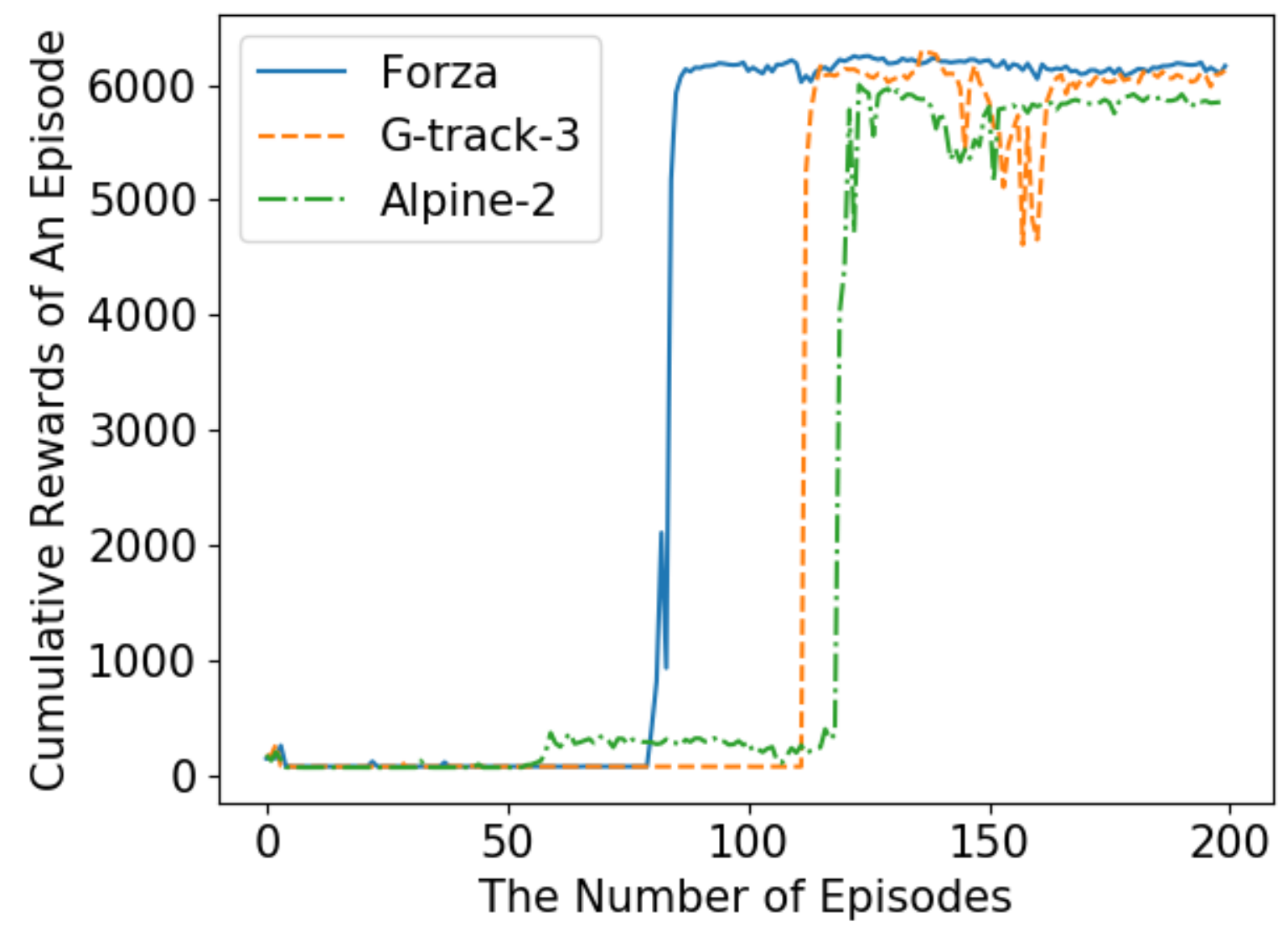}
	\caption{The cumulative reward curve of the training process. The agent converges faster in the easier track forza than the difficult track alpine-2.}
	\label{fig_return}
\end{figure}

We train the agent in three different tracks: forza, g-track-3 and alpine-2 as shown in Fig. \ref{fig_data_collector}. These three tracks are in different levels of difficulty. The easiest forza is mostly made of straight roads, and the most difficult alpine-2 is mostly made of shape turns, uphill and downhill slopes. The training rewards of different tracks are illustrated in Fig. \ref{fig_return}. At the beginning phase, the agent is on the straight segment and steers poorly. It always keeps turning right or left until out of track. After a period of exploration, it learns from history and gradually is able to conduct proper steering to control the vehicle in the lane. Finally, the neural network converges to a decent steering policy in all three tracks. An interesting fact is that the learning speed varies among different tracks. In the forza track, the road is mostly made of straight segment, so the agent only needs to learn how to drive in the straight road and mild curve. The g-track-3 is a little harder than forza because a large radius curve follows the straight road. Therefore, the agent must learn how to drive at the straight road and the curve. The alpine-2 is much harder because it contains straight road, sharp curve and uphill/downhill slope. Thus, the agent learns slowly at this track. The forza track's final cumulative reward is slightly the highest, which is consistent with the track difficulty.

\begin{table*}[!t]
	\renewcommand{\arraystretch}{1.4}
	\setlength{\tabcolsep}{1.4em}
	\centering
	\caption{Lateral Controllers Performance Comparison}
	\label{pid_lqr_rl_cmp}
	\begin{tabular}{c|c|c|c|c|c|c|c|c|c}
		\hline \hline
		\,  & \multicolumn{5}{|c|}{LQR Setup} & MPC Setup & \multicolumn{3}{|c}{Scores} \\
		\cline{2-10}
		Tracks & $ q_1 $ & $ q_2 $ & $ q_3 $ & $ q_4 $ & $ \rho $ & $H_p$ & LQR & MPC & RL\\
		\hline
		Forza & 2.0 & 1.0 & 2.0 & 0.2 & 0.05 & 8  & 6333.1 & 6348.1 & \textbf{6372.3} \\
		Forza & 2.0 & 0.2 & 2.0 & 0.1 & 0.01 & 10 & 6335.5 & 6346.3 & \textbf{6375.1} \\
		Forza & 1.0 & 0.2 & 1.0 & 0.1 & 0.01 & 12 & 6335.9 & 6344.7 & \textbf{6372.9} \\
		\hline
		Alpine-2 & 2.0 & 1.0 & 2.0 & 0 & 0.05 & 8  & 4400.0 & 4411.6 & \textbf{4415.6} \\
		Alpine-2 & 2.0 & 0.3 & 2.0 & 0 & 0.01 & 10 & 4364.0 & 4405.4 & \textbf{4419.4} \\
		Alpine-2 & 2.0 & 0.5 & 1.0 & 0 & 0.01 & 12 & 4400.1 & 4401.2 & \textbf{4415.9} \\
		\hline
		Eroad & 3.0 & 0.2 & 1.5 & 0 & 0.03 & 8 & 3585.9 & \textbf{3603.6} & 3592.8 \\
		Eroad & 1.0 & 0.8 & 2.5 & 0 & 0.01 & 10 & \textbf{3591.9} & 3589.0 & \textbf{3593.9} \\
		Eroad & 1.5 & 0.5 & 1.5 & 0.03 & 0.05 & 12 & 3520.9 & 3557.4 & \textbf{3592.9} \\
		\hline
		G-track-3 & 2.0 & 1.0 & 2.0 & 1.0 & 0.05 & 8  & 3110.3 & \textbf{3210.5} & \textbf{3213.5}\\
		G-track-3 & 2.0 & 0.2 & 2.0 & 0.1 & 0.01 & 10 & 3209.4 & 3206.3 & \textbf{3212.4} \\
		G-track-3 & 1.0 & 0.2 & 1.0 & 0.1 & 0.01 & 12 & 3184.6 & 3187.1 & \textbf{3215.3} \\
		\hline
		\hline
	\end{tabular}
\end{table*}

\subsection{Comparison with Other Methods}
The traditional control methods like LQR and MPC play an essential role both in simulation and in real vehicle steering control. Here we want to answer the question that whether the RL agent can learn a better policy than the LQR and MPC.

\subsubsection{LQR controller}
The objective is to control the vehicle running along the track center. Therefore, the errors can be defined as:
\begin{subequations}
	\begin{equation}\label{pid_e1}
		e_1(t) = d(t) - d_0(t),
	\end{equation}
	\begin{equation}\label{pid_e2}
		e_2(t) = \psi(t) - \psi_0(t),
	\end{equation}
\end{subequations}
where $ e_1(t) $ is distance error and $ e_2(t) $ is angle error at time $ t $. Here the system input $d(t)$ and $\psi(t)$ are measured from the range meter and angle sensor, respectively. The range meter gives the distance to lane center and the angle sensor gives the yaw angle between vehicle and lane orientation. The reference distance to track center $ d_0(t) = 0 $ and reference angle $ \psi_0(t) $ is track heading angle.

Define the state variable as $ \mathbf{x}=[e_1(t), \dot{e}_1(t), e_2(t), \dot{e}_2(t)]^{\intercal} $. Following the vehicle dynamic model in terms of errors in \cite{Rajamani2012}, the state space model is given by
\begin{equation}\label{lqr_model}
\dot{\mathbf{x}} = A\mathbf{x} + B\delta(t),
\end{equation}
where $ \delta(t) $ is front wheel steering angle. The system dynamics is defined as:
\begin{subequations}
	\begin{equation}\label{lqr_a}
		A = \left[
		\begin{array}{cccc}
		0 & 1 & 0 & 0 \\
		0 & -\frac{2C_f + 2C_r}{mv_x} & \frac{2C_f + 2C_r}{m} & \frac{-2C_f l_f + 2C_r l_r}{mv_x} \\
		0 & 0 & 0 & 1 \\
		0 & \frac{-2C_f l_f + 2C_r l_r}{I_zv_x} & \frac{2C_f l_f - 2C_r l_r}{I_z} & \frac{-2C_f l_f^2 + 2C_r l_r^2}{I_z v_x}
		\end{array}
		\right],
	\end{equation}
	\begin{equation}
		B = \left[
		\begin{array}{c}
		0 \\
		\frac{2C_f}{m} \\
		0 \\
		\frac{2C_f l_f}{I_z}
		\end{array}
		\right].
	\end{equation}
\end{subequations}
The parameters $ C_f $ and $ C_r $ are cornering stiffness of front and rear tires. $ l_f $ and $ l_r $ are longitudinal distances from center of gravity to the front and rear tires. $ m $ is vehicle mass and $ I_z $ is yaw moment of inertia. $ v_x $ is speed along vehicle heading direction. We choose cost matrix for state and action as:
\begin{subequations}
	\begin{equation}\label{lqr_q}
	Q = \left[
	\begin{array}{cccc}
	q_1 & 0 & 0 & 0 \\
	0 & q_2 & 0 & 0 \\
	0 & 0 & q_3 & 0 \\
	0 & 0 & 0 & q_4
	\end{array}
	\right],
	\end{equation}
	\begin{equation}\label{lqr_r}
	R = \rho I,
	\end{equation}
\end{subequations}
where $ q_1 \sim q_4 $ and $ \rho $ are cost coefficients to be tuned.

We use the TORCS vehicle car1-trb1 to implement all experiments. The model parameters for the vehicle are as follows: $ C_f = 80\textrm{,}000$, $ C_r = 80\textrm{,}000 $, $ l_f = 1.27 $, $ l_r = 1.37 $, $ m = 1\textrm{,}150 $, and $ I_z = 2\textrm{,}000 $.

\subsubsection{MPC controller}
The MPC controls the front-wheel steering angle $\delta$ by using the following nonlinear kinematic formulations according to\cite{Kong2015kinematic}:
\begin{subequations}\label{nmpc_eq}
	\begin{equation}\label{nmpc_eq1}
	\dot{x} = v\cos(\psi + \beta),
	\end{equation}
	\begin{equation}\label{nmpc_eq2}
	\dot{y} = v\sin(\psi + \beta),
	\end{equation}
	\begin{equation}\label{nmpc_eq3}
	\dot{\psi} = \frac{v\cos(\beta)}{l_f + l_r}\tan(\delta),
	\end{equation}
	\begin{equation}
	\beta = \tan^{-1}(\frac{l_r}{l_f + l_r}\tan(\delta)),
	\end{equation}
\end{subequations}
where $x$ and $y$ are the coordinates of the center of gravity in an inertial frame $(X, Y)$. $\psi$ is the orientation angle of vehicle with respect to axis $X$, $v$ is the vehicle speed and $\beta$ is the slip angle. Define the state $\xi=[x, y, \psi]^{\intercal}$ and action $u = \delta \in \mathcal{U}$, then by according to (\ref{nmpc_eq}), the nonlinear kinematic vehicle model can be defined as 
\begin{subequations}\label{nonlinear_model}
	\begin{equation}
		\dot{\xi} = f^{dt}(\xi, u),
	\end{equation}
	\begin{equation}
		\zeta = h(\xi),
	\end{equation}
\end{subequations}
where the output $\eta$ is given by
\begin{equation}\label{nonlinear_model_output}
h(\xi) = 
\left[
\begin{matrix}
0 & 1 & 0 \\
0 & 0 & 1
\end{matrix}
\right]\xi.
\end{equation}
The vehicle model is discretized by using Euler methods and sampling time $dt = 50 $ ms. Given the reference output $\zeta_{ref}=[y_{ref}, \psi_{ref}]^{\intercal}$ and prediction horizon $H_p$, the MPC controller aims to solve the following constrained finite-time optimal control problem:
\begin{equation}
	\begin{aligned}
	& \underset{\mathcal{U}}{\text{min}}
	& & \sum_{i=1}^{H_p} (\zeta_i - \zeta_{ref, i})^{\intercal}Q(\zeta_i - \zeta_{ref, i})  + \sum_{i=1}^{H_p - 1}  + u_i^{\intercal}R u_i, \\
	& \text{s.t.} & & \xi_{k+1} = f^{dt}(\xi_k, u_k), k = 1,...,H_p, \\
	& & & \zeta_{k} = h(\xi_k), k = 1,...,H_p, \\
	& & & \delta_{min} \le u_k \le \delta_{max}, k = 1,...,H_p-1, \\
	\end{aligned}
\end{equation}
where $Q$ and $R$ are weighting diagonal matrices of appropriate dimensions. We choose all weights as 1.0.

\subsubsection{RL controller}
The RL controller is the neural network trained above. The state variable is composited by the measurements from the range meter, angle sensor, and speed meter. These sensors measure the distance to lane center, yaw angle between vehicle and lane orientation, and the vehicle speed, respectively. In order to validate RL controller's generalization capability across different tracks, we use the controller trained in g-track-3 track to implement all experiments.

In all experiments, three controllers are employed to control the same vehicle at the same track. The coefficients of LQR and prediction horizon of MPC are explored in Table \ref{pid_lqr_rl_cmp}. The reference trajectory of the MPC is the track center line, and $8\textrm{-}12$ prediction horizon covers about $35\textrm{-}50$ meters. The score computation follows (\ref{eq_reward}). We totally select four typical tracks as shown in Fig. \ref{fig_data_collector}: a) forza, which is a high-speed track mostly formed by straight segment, b) alpine-2, which is emphasis on elevation changes, c) eroad, which includes a rich set of curves, and d) g-track-3, which is a good choice for balancing straight road, curve, and elevation changes. Table \ref{pid_lqr_rl_cmp} shows that the RL controller slightly outperforms the best LQR controllers in all tracks. The RL agent can also perform better than nonlinear MPC controller in some tracks, e.g. forza and alpine-2. The better performances in these tracks validate that the model-free RL controller can learn a decent policy by interacting with the environment. An interesting fact is that if we fix the same coefficient set for the LQR controller, it can hardly generalize to new tracks like g-track-3 and alpine-2. We also train a RL controller with quadratic reward $r=-(0.3e_1(t)^2+e_2(t)^2+0.03\delta(t)^2)$. The performance of the quadratic is comparable to the reward in (\ref{eq_reward}). For example, the new RL controller achieves 6373.9 in forza.

\subsection{Vision-based Lateral Control}
Now we implement visual lateral controller by utilizing the proposed MTL-RL controller. As shown in Fig. \ref{fig_sys_framework}, first the MTL perception network receives a driver-view image $ o_t $ and predicts the underlying track feature vector $ \sigma_t $. Then together with the vehicle property vector $\eta_t$, the RL controller computes the steering command $ a_t $ based on state $ s_t = [\sigma_t, \eta_t] $. Finally, VTORCS executes the steering command $ a_t $.

\begin{figure}[!t]
	\centering
	\includegraphics[scale=0.535]{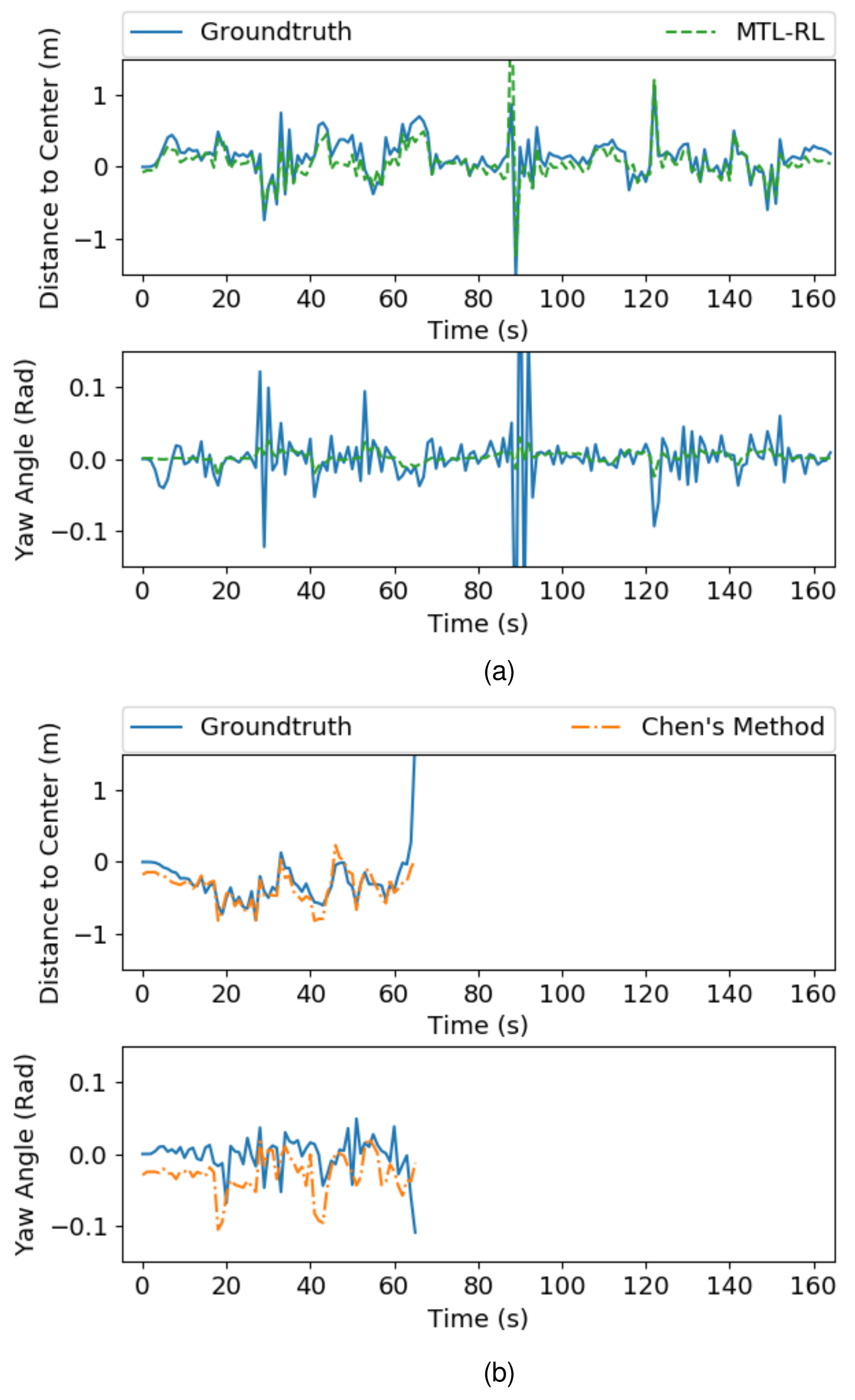}
	\caption{The (a) MTL-RL controller and (b) Chen's method driving curves of ground-truth and prediction for distance to middle and heading angle in g-track-3. With visual input, the proposed MTL-RL controller can finish a lap, while Chen's method cannot.}
	\label{fig_mtl_drive}
\end{figure}

We test the MTL-RL controller in g-track-3. The distance to track center and yaw angle are shown in Fig. \ref{fig_mtl_drive} (a). For comparison, we also plot the ground-truth. It is seen from the figure that the predicted values generally match the ground-truth. Since the MTL perception network can predict accurate state $s$, the vehicle achieves the score 3175.9 with only visual input. This score is comparable to the scores of the LQR (3209.4) controller with the accurate physical measurement as its input. Fig. \ref{fig_mtl_drive} (a) shows that the MTL-RL controller is capable of controlling the maximal distance prediction errors within 0.4m. Note that for the discrepancy of the predicted results and the ground-truth between $t=80s$ and $t=100s$ in Fig. \ref{fig_mtl_drive}, it is caused by the curve and heavy shadow. The perception errors cause the vehicle jittering in this period and leading to the discrepancy. However, after a few seconds, the RL controller stabilizes the vehicle in which case the distance to lane center and yaw angle are near zero. We also compare the performance with Chen's \cite{chen2015deepdriving} perception model whose results are 
shown in Fig. \ref{fig_mtl_drive} (b). The image is fed to CNN model to predict distance to lane center and yaw angle. The same RL controller as the MTL-RL experiment takes the state $ s_t $ and makes the steering command $ a_t $. The experiment is also implemented in g-track-3 for fair comparison. The vehicle runs out of the track at the 66th second because the perception network works poorly at a shadowy sharp curve. 

\begin{figure}[!t]
	\centering
	\includegraphics[scale=0.53]{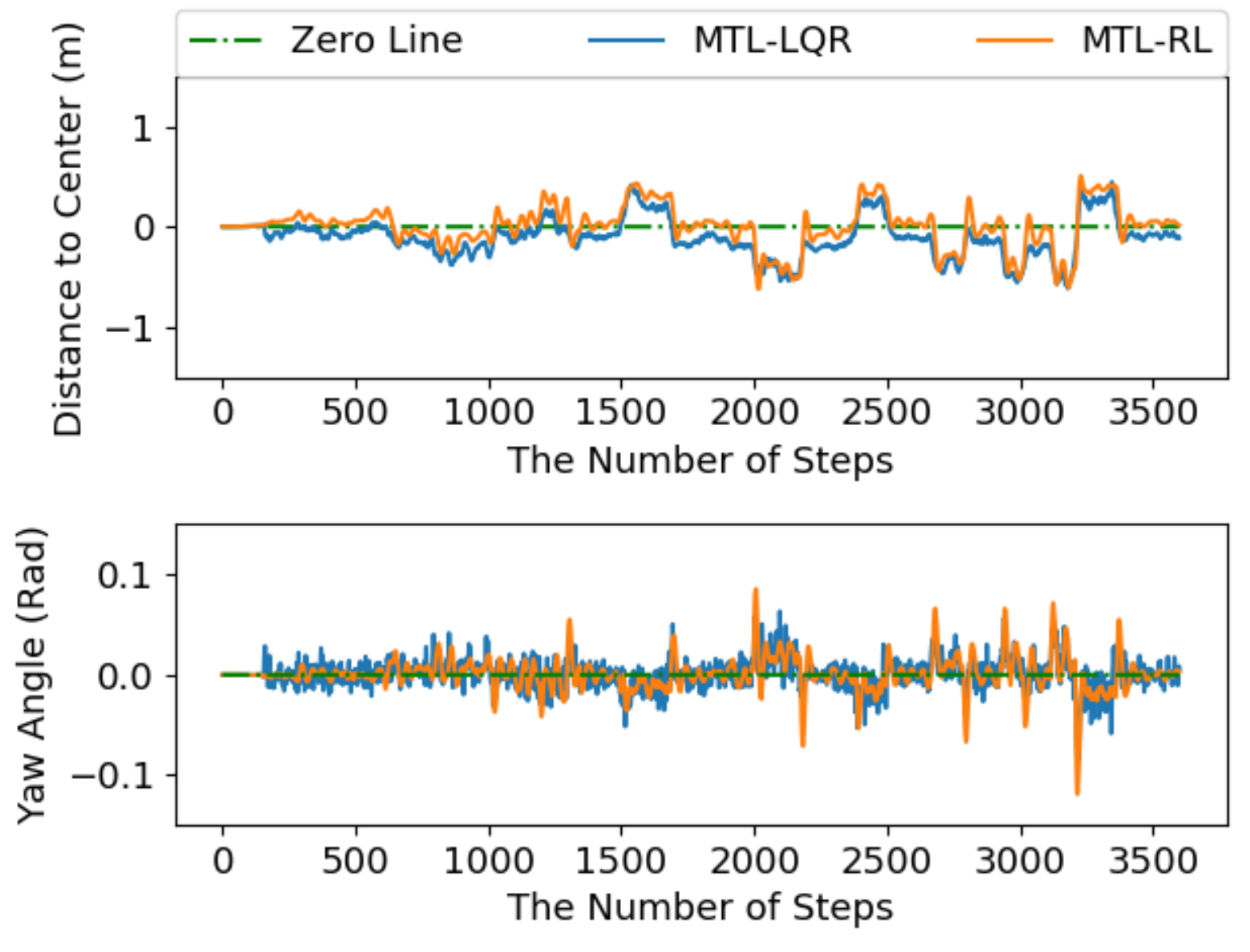}
	\caption{The distance to middle and yaw angle of MTL-RL and MTL-LQR controllers in alpine-2. With visual input, both controllers can successfully control the vehicle along lane center, but MTL-RL is more robust to the perception noise in average.}
	\label{fig_mtl-rl_mtl-lqr_cmp}
\end{figure}

Additionally, we compare the performance of the MTL-RL controller with MTL-LQR controller whose input vector is also built from MTL perception features in alpine-2 track. The cost coefficients of LQR controller are chosen as $(q_1,q_2,q_3,q_4,\rho) = (2, 0.5, 1.0, 0, 0.01)$ which is the coefficients corresponding to the highest score in alpine-2 track. The distance to lane center and yaw angle are shown in Fig. \ref{fig_mtl-rl_mtl-lqr_cmp}. Both of the MTL-RL and MTL-LQR can successfully control the vehicle to run along lane center with driver-view image as the input. The average distances to lane center of MTL-RL and MTL-LQR controllers are 0.148 m and 0.175 m, respectively. And the average yaw angle of MTL-RL and MTL-LQR controllers are both 0.01 radian. Since the MTL perception features are noisy compared with the ground-truth, the MTL-RL controller is more robust to the noise than MTL-LQR controller in average.

\section{Conclusion}
In this paper, we propose a framework for vision-based lateral control which combines DL and RL methods. In order to improve the perception accuracy, an MTL CNN model is proposed to learn the key track features, which are used to locate the vehicle in the track coordinate. A policy gradient RL controller is trained to solve the continuous sequential decision-making problem. By combining the MTL perception module and RL control module, the MTL-RL controller is capable of controlling the vehicle run along the track center with driver-view image as its input. Additionally, we propose the VTORCS environment based on TORCS. It helps to provide high-quality image stream and easy-to-use RL interface, which enables efficient data collection and algorithm implementation.

The experiments show that the MTL perception network can stably and accurately predict the track features and locate the vehicle in the track coordinate. It validates that learning multiple track perception tasks jointly leads to a lower test error than the single task learning. The trained RL controller shows good generalization across various tracks with different difficulties. It outperforms the popular control methods like LQR in all testing tracks. 

Since the fully autonomous driving control includes both the lateral control and the longitudinal control, one future direction is to combine the lateral and the longitudinal control. In the fully autonomous controller, one of the important aspects is the adaptation to different driver's styles, e.g. preferred maximum acceleration and headway time in longitudinal control. How to model the driving styles of different drivers and incorporate them in the controller design are challenging problems. One possible way is to classify the driving styles and fuse the style factor in the reward design. Additionally, the learning method is now the model-free method which causes long learning process. To accelerate learning, we intend to combine the model-free and the model-based learning methods by approximating the system dynamics in a small time interval. Last, the perception input now is the static image, but the driving images have strong correlation features among continuous frames. These temporal features, for example road orientation change and velocity to lane markings can give more information to the RL agent to facilitate the decision-making. Therefore, a recurrent neural network (RNN) may be built on the top of MTL perception model to extract the temporal features.


%

\section*{Acknowledgment}
This work is supported by the Beijing Science and Technology Plan under Grants No. Z181100004618003 and No. Z181100008818075, the National Natural Science Foundation of China (NSFC) under Grants No. 61573353, No. 61803371, No. 61573353, and No. 61533017, the National Key Research and Development Plan under Grant No. 2016YFB0101003.

\ifCLASSOPTIONcaptionsoff
  \newpage
\fi



\bibliographystyle{IEEEtran}
\bibliography{IEEEabrv,IEEEexample}
%


%




\end{document}